\documentclass{article}

\usepackage[preprint]{corl_2026}

\usepackage{amsmath}
\usepackage{amssymb}
\usepackage{bm}
\usepackage{hyperref}
\usepackage{cleveref}

\usepackage{graphicx}
\usepackage{booktabs}
\usepackage{multirow}
\usepackage{makecell}
\usepackage{subcaption}
\usepackage{enumitem}
\usepackage{float}
\usepackage{algorithm}
\usepackage{algpseudocode}

\title{Learning Unions of Convex Sets via Invertible Latent Decomposition for Path Planning}

\author{
  Taerim Yoon$^{1}$\\
  Korea University\\
  \texttt{taerimyoon@korea.ac.kr}
  \And
  Dongho Kang$^{2}$\\
  ETH Zurich\\
  \texttt{kangd@ethz.ch}
  \And
  Kisang Park$^{1}$\\
  Korea University\\
  \texttt{kisangpark@korea.ac.kr}
  \AND
  Junha Cha$^{1}$\\
  Korea University\\
  \texttt{junhacha@korea.ac.kr}
  \And
  Stelian Coros$^{2}$\\
  ETH Zurich\\
  \texttt{scoros@ethz.ch}
  \And
  Sungjoon Choi$^{1}$\\
  Korea University\\
  \texttt{sungjoon-choi@korea.ac.kr}
}

\begin{document}
\maketitle

\begin{abstract}
Collision-free path planning in cluttered, real-world environments relies on a representation
of the collision-free space, and existing representations broadly fall into two categories.
\textit{Explicit} representations, such as unions of convex sets, can be plugged into optimization-based planners as hard collision-free constraints, but their parameters scale poorly with configuration-space dimension.
\textit{Implicit} representations, by contrast, are flexible and scale well to complex geometries, yet typically lack such guarantees.
We bridge this gap with \textbf{ILD} (\textbf{I}nvertible \textbf{L}atent \textbf{D}ecomposition), a framework that jointly learns an invertible mapping and a union of explicit convex polytopes in the resulting latent space.
Planning is carried out over these latent convex sets, and the invertible mapping decodes the resulting paths back to the original configuration space while preserving feasibility with respect to the refined explicit safe regions.
We further propose Visibility-Guided Sampling (VGS) to keep the convex sets connected for path planning.
Across 2D navigation, 6-DoF, and 14-DoF manipulation environments, ILD achieves broader coverage, better inter-set connectivity, and higher path-planning success rates than prior baselines, with zero observed false positives after test-time refinement.
On a 14-DoF bimanual manipulator, we further demonstrate real-time collision-free planning, with test-time refinement adapting to scene-geometry changes during real-world deployment on a single 6-DoF arm.
\end{abstract}

\keywords{Combination of learning and planning in robotics, Robot safety, alignment, and safe learning-based systems, Latent Convex Decomposition, Invertible Neural Networks}

\section{Introduction}

Representing the collision-free region of a robot's configuration space is a core building block which enables collision-free planning for a wide range of tasks including navigation and manipulation.
Such representations broadly fall into two categories, depending on whether the region is described \textit{explicitly} or \textit{implicitly}.
\textit{Explicit representations} characterize the collision-free region through geometric primitives whose parameters are directly accessible, such as the convex sets consisting of hyperplanes~\citep{deits2015computing, marcucci2023motion, drake} or the graph vertices and edges produced by sampling-based planners~\citep{kuffner2000rrtconnect, kavraki1996prm}.
This explicitness allows the representation to be plugged into planners as hard collision-free constraints, but the parameters required for broad coverage grow rapidly with the dimensionality of the configuration space, limiting scalability~\citep{werner2024approximating, werner2024faster}.
\textit{Implicit representations} instead encode the collision boundary as the level set of a learned function, offering greater flexibility for complex geometries~\citep{koptev2022neural, guo2025deepcollide}.
However, because the collision-free region is defined only implicitly, it cannot
be incorporated as hard constraints in optimization-based planners, and therefore provides no explicit collision-free constraints.

To bridge these two directions, we aim to learn a union of convex polytopes in a structured latent space, combining the data-driven flexibility of learned representations with the hard-constraint structure required for explicit collision-free constraints.
Each polytope is described by a small set of linear half-space constraints, so the resulting representation can be passed to an optimization-based planner as explicit hard constraints.
The central challenge in this approach is that a region learned as collision-free in a latent space does not necessarily correspond to a collision-free region in the original configuration space.
Our key insight is to learn this mapping with an invertible neural network, whose structural invertibility guarantees a one-to-one correspondence between the configuration space and the latent space. Therefore, any latent region verified or refined to be collision-free can be mapped back consistently to the original configuration space.
To this end, we propose \textbf{ILD} (\textbf{I}nvertible \textbf{L}atent \textbf{D}ecomposition), which jointly learns the invertible mapping together with a union of convex polytopes in the resulting latent space.
Crucially, the latent space is jointly shaped to be well-suited for convex covering, so the collision-free region admits a far more compact decomposition than in the original space, reducing both representation size and the cost of downstream planning.
Since this mapping is invertible, plans optimized over the explicit convex sets in the latent space, for instance with the Graph of Convex Sets (GCS) framework~\citep{marcucci2023motion}, can be decoded back to the original configuration space while preserving feasibility with respect to the explicit safe regions.
Around this core, we add two practical components. The first is Visibility-Guided Sampling (VGS), which encourages inter-region connectivity. The second is a test-time refinement procedure that corrects false positives at deployment.

We evaluate the framework on 2D navigation, 6-DoF, and 14-DoF manipulation environments.
The proposed method achieves broader collision-free region coverage, improves connectivity between convex sets, and increases path-planning success rates compared to baseline methods.
Notably, after test-time refinement, the model attains precision $=1.0$ on the evaluated samples and on the false positives observed during deployment, without retraining.

Our contributions are summarized as follows:
\begin{itemize} [leftmargin=1.0em,itemsep=0.15em,topsep=0.2em]
    \item We propose ILD, which learns a compact union of convex polytopes in an invertible latent space, with broad coverage and zero observed false positives after test-time refinement.
    \item VGS improves connectivity between convex sets, which results in higher planning success rates on optimization-based planners such as the Graph of Convex Sets (GCS) planner~\citep{marcucci2023motion}.
    \item We validate our approach on a 14-DoF bimanual manipulator, with deployment-time refinement enabling adaptation to scene-geometry changes on a single 6-DoF arm of the real-world platform without retraining.
\end{itemize}

\section{Related Work}

\subsection{Explicit Representations}

\noindent\textbf{Sampling-based methods}
A classical family of explicit methods builds a discrete graph in the collision-free region.
Sampling-based planners such as Rapidly-exploring Random Tree (RRT)~\citep{kuffner2000rrtconnect} and Probabilistic Roadmap (PRM)~\citep{kavraki1996prm} populate this graph with verified configurations as vertices and locally checked transitions as edges, yielding paths that are collision-free by construction.
GPU-parallel variants such as cuRobo~\citep{sundaralingam2023curobo} further amortize collision checks for fast per-query planning.
However, since the collision-free space is characterized only as a discrete set of configurations and edges, the sample density needed for adequate coverage scales exponentially with the configuration-space dimension, a major bottleneck for high-DoF systems.

\noindent\textbf{Region-based methods}
An alternative line of work represents the collision-free space as a union of convex sets, which can be integrated directly into optimization-based frameworks such as motion planning around obstacles with Graphs of Convex Sets (GCS)~\citep{marcucci2023motion, drake}. Notably, algorithms like Iterative Regional Inflation by Semidefinite Programming (IRIS)~\citep{deits2015computing} and its accelerated variants construct collision-free convex polytopes by alternating between finding separating hyperplanes and maximizing inscribed ellipsoids. To mitigate the complexity of these representations, recent approaches such as that by \citet{werner2024approximating} reduce the total number of required regions by computing clique covers of visibility graphs.
Unlike sampling-based methods, these approaches expose the collision-free region at a volumetric level rather than through discrete samples, allowing optimization-based planners to enforce safety as continuous hard constraints over an entire region rather than routing between verified edges.
However, the number of required convex sets still scales poorly with dimensionality, inflating both the per-region construction cost and the downstream planner's search complexity.
While ILD shares the objective of generating explicit convex sets for optimization-based planners and inherits their explicit collision-free constraints, it shifts the construction process to a latent space by jointly learning the convex decomposition alongside an invertible mapping.

\subsection{Implicit Representations}

Another line of work models the collision-free region as an implicit function learned from data, enabling it to scale to more complex configurations and environments.
\citet{koptev2022neural} learn neural signed distance functions in joint configuration space, enabling reactive control.
\citet{guo2025deepcollide} learn an implicit collision detection function from sampled configuration-label pairs, using a forward kinematics kernel and Fourier positional encoding to scale to high-DoF settings with inference time independent of training set size.
While these approaches offer flexibility and scale effectively to complex geometries, they do not provide explicit collision-free constraints along decoded paths.
ILD operates in the same data-driven setting, but produces explicit convex sets that can be directly used as hard constraints by optimization-based planners.

\section{Methods}

\begin{figure*}[t]
    \centering
    \includegraphics[width=0.99\linewidth]{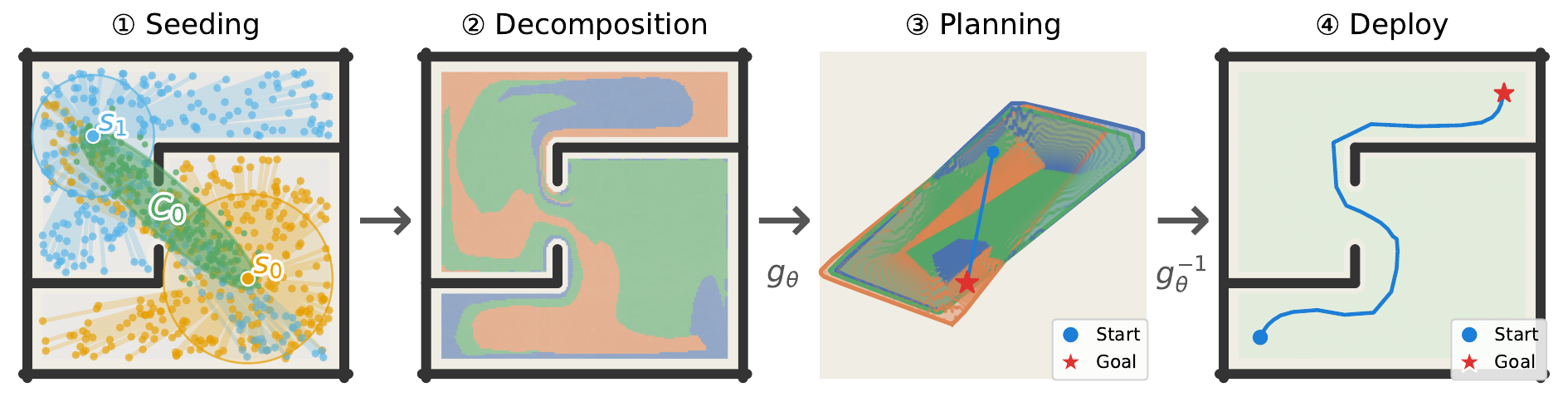}
    \caption{
        Overview of ILD on a 2D navigation task. ILD proceeds in four stages.
        \textbf{(1) Seeding:} This example places two seeds $s_0, s_1$ and one bridge $c_0$ connecting them. Each seed sampling region $\mathcal{S}_i$ covers the configurations visible from $s_i$ (orange, blue), and the bridge sampling region $\mathcal{C}_0$ covers a neighborhood around $c_0$ (green ellipse).
        \textbf{(2) Decomposition:} We jointly optimize $g_\theta$ with the explicit convex sets to cover each region in the latent space, partitioning $\mathcal{Q}_\text{free}$ into colored regions in configuration space.
        \textbf{(3) Planning:} In the latent space $\mathcal{Z}$, a Graph of Convex Sets (GCS) solver~\citep{marcucci2023motion, drake} returns a collision-free path through the convex sets, which we decode back to $\mathcal{Q}$ via $g_\theta^{-1}$.
        \textbf{(4) Deploy:} The decoded trajectory is executed on the robot, with collision-free guarantees inherited from the latent path by invertibility of $g_\theta$.
    }
    \label{fig:method_overview}
    \vspace{-16pt}
\end{figure*}

Let $\mathcal{Q} \subset \mathbb{R}^n$ denote the robot's configuration space and $\mathcal{Q}_\text{free} \subseteq \mathcal{Q}$ its collision-free subset.
ILD learns an invertible mapping that deforms $\mathcal{Q}$ into a latent space where $\mathcal{Q}_\text{free}$ can be compactly covered by a union of convex polytopes.
Invertibility guarantees that any path planned in the latent space decodes back to a valid, collision-free trajectory in the original space.
We describe the latent decomposition and its training in Section~\ref{sec:lcd}.
To encourage inter-region connectivity, we introduce a Visibility-Guided Sampling (VGS) strategy in Section~\ref{sec:vgs}, which is critical for path planning across the resulting decomposition.
Finally, since the explicit linear boundaries allow local correction without retraining, we describe a test-time refinement procedure for handling false positives in Section~\ref{sec:refinement}.

\subsection{Problem Formulation}

Our goal is to jointly learn an invertible mapping $g_\theta : \mathcal{Q} \to \mathcal{Z}$ and a union of convex polytopes $\{\mathcal{P}_k\}_{k=1}^{N}$ in the latent space $\mathcal{Z}$ such that $\hat{\mathcal{Q}}_\text{free} = g_\theta^{-1}(\bigcup_{k=1}^{N} \mathcal{P}_k)$ covers $\mathcal{Q}_\text{free}$, where each polytope $\mathcal{P}_k = \{ \mathbf{z} \in \mathcal{Z} \mid \boldsymbol{\eta}_{k,i}^\top \mathbf{z} + d_{k,i} \geq 0,\; i = 1, \dots, B \}$ is defined by linear half-space constraints in the latent space. The invertibility of $g_\theta$ ensures that $\bigcup_k \mathcal{P}_k$ and $\hat{\mathcal{Q}}_\text{free}$ are in one-to-one correspondence.
We aim for two objectives, \textit{zero false positives} (precision\,=\,1) and \textit{broad coverage} (high recall), while using as few convex sets $N$ as possible.
Since an optimization-based planner over convex sets can only route paths within a single connected component, we maximize coverage restricted to the largest connected component of $\hat{\mathcal{Q}}_\text{free}$.
These convex sets are then used directly as hard constraints by any planner that consumes a union of convex polytopes. In our experiments we instantiate this planner with the Graph of Convex Sets (GCS) framework~\citep{marcucci2023motion, drake}.

\subsection{Latent Convex Decomposition}
\label{sec:lcd}

\paragraph{Invertible latent mapping.}
Covering $\mathcal{Q}_\text{free}$ using a union of convex polytopes is challenging, as $\mathcal{Q}_\text{free}$ is typically non-convex, especially in high-DoF configuration spaces.
To address this challenge, we introduce an invertible mapping $g_\theta : \mathcal{Q} \to \mathcal{Z}$ that warps the configuration space into a latent space $\mathcal{Z}$ where $\mathcal{Q}_\text{free}$ admits a compact decomposition into a few convex polytopes.
We implement $g_\theta$ as a stack of invertible linear layers with LU decomposition~\citep{kingma2018glow} interleaved with invertible nonlinear activations, with architectural details in Appendix~\ref{app:architecture}.

\paragraph{Region representation.}
For a latent point $\mathbf{z} = g_\theta(\mathbf{q})$, region $k$ is defined by $B$ hyperplanes with learnable parameters $\boldsymbol{\eta}_{k,i}\in\mathbb{R}^n$, $d_{k,i}\in\mathbb{R}$, and $\mathbf{z}$ is inside it iff all signed distances $\phi_{k,i}(\mathbf{z}) = \boldsymbol{\eta}_{k,i}^\top \mathbf{z} + d_{k,i}$ are non-negative.
We aggregate per-region membership and the union over regions by smooth Gumbel-Softmax surrogates of $\min$ and $\max$:
\begin{equation}
    \varphi_k(\mathbf{z}) \approx \min_i\, \phi_{k,i}(\mathbf{z}), \qquad
    C(\mathbf{z}) \approx \sigma\!\left(\max_k\, \varphi_k(\mathbf{z})\right),
\end{equation}
where $\varphi_k>0$ marks $\mathbf{z}$ as inside region $k$ and $C(\mathbf{z})\ge 0.5$ marks it as inside the union. We write $C_k(\mathbf{z}) = \sigma(\varphi_k(\mathbf{z}))$ for the per-region membership probability.
We adopt the hyperplane parameterization of CvxNet~\citep{deng2020cvxnet}, originally proposed for 3D shape modeling, and apply it to higher-dimensional configuration spaces ($n > 3$).

\paragraph{Joint training objective.}
We jointly optimize the latent mapping $g_\theta$ and the boundary parameters $\mathcal{H} = \{\boldsymbol{\eta}_{k,i}, d_{k,i}\}_{k=1,\,i=1}^{N,\,B}$ by minimizing the classification loss $\mathcal{L}_\text{class}$ together with the per-seed, per-bridge, and regularization terms:
\begin{equation}
    \hat{\theta}, \hat{\mathcal{H}} = \arg\min_{\theta,\, \mathcal{H}}\; \mathcal{L}_\text{class} + \lambda_\text{seed}\, \mathcal{L}_\text{seed} + \lambda_\text{vis}\, \mathcal{L}_\text{vis} + \lambda_\text{bridge}\, \mathcal{L}_\text{bridge} + \lambda_\text{reg}\, \mathcal{L}_\text{reg},
\end{equation}
where $\mathcal{L}_\text{class} = \mathrm{BCE}(C_\text{gt},\, C(g_\theta(\mathbf{q});\, \mathcal{H}))$ is the per-sample binary cross-entropy between predicted probabilities and ground-truth collision labels. Geometrically, this objective encourages $g_\theta$ to warp non-convex parts of $\mathcal{Q}_\text{free}$ to fit the convex polytopes (Appendix~\ref{app:bce_gradient}).
The per-seed, per-visible, and per-bridge losses $\mathcal{L}_\text{seed},\, \mathcal{L}_\text{vis},\, \mathcal{L}_\text{bridge}$ are introduced by the VGS strategy in Section~\ref{sec:vgs}, and the regularization losses $\mathcal{L}_\text{reg}$, including an isometric loss $\mathcal{L}_\text{iso} = \mathbb{E}[(\|g_\theta(\mathbf{q}_1) - g_\theta(\mathbf{q}_2)\| - \|\mathbf{q}_1 - \mathbf{q}_2\|)^2]$ so that latent shortest paths stay shortest in $\mathcal{Q}$, are described in Appendix~\ref{app:reg_losses}.

\subsection{Visibility-Guided Sampling}
\label{sec:vgs}

Convex decomposition methods such as IRIS~\citep{werner2024faster} and CvxNet~\citep{deng2020cvxnet} grow each region around a \textit{seed} configuration, and the choice of seeds critically shapes the resulting safe region.
We choose seeds based on visibility to improve coverage and connectivity, where two configurations are \textit{visible} to each other if the straight-line segment between them is collision-free. The intuition is that since $g_\theta$ can flexibly deform $\mathcal{Q}$, a set of mutually visible configurations in $\mathcal{Q}_\text{free}$ can be mapped to a single convex polytope $\mathcal{P}_k$ in the latent space, even when they form a non-convex region in $\mathcal{Q}$. For example, in Figure~\ref{fig:method_overview}, $g_\theta$ maps the blue visibility set of seed $s_1$ to a single convex polytope in the latent space.
In detail, VGS consists of \textit{seed sampling} and \textit{bridge sampling}, described below.

\paragraph{Seed sampling.}
Each seed sampling region aims to cover a large set of mutually visible configurations as a single convex polytope and thereby maximize coverage. We start by constructing a visibility matrix over uniformly sampled candidate configurations. We then greedily pick $N_\text{seed}$ seeds with the most visible points (Figure~\ref{fig:method_overview}, step (1), with selection details in Appendix~\ref{app:seed_selection}). For each chosen seed $s_k$, we draw $\mathbf{q}_\text{seed}$ from a neighborhood of $s_k$ and $\mathbf{q}_\text{vis}$ from its visible candidates, yielding two per-seed BCE terms $\mathcal{L}_\text{seed} = \mathrm{BCE}(C_\text{gt},\, C_k(g_\theta(\mathbf{q}_\text{seed})))$ and $\mathcal{L}_\text{vis} = \mathrm{BCE}(C_\text{gt},\, C_k(g_\theta(\mathbf{q}_\text{vis})))$.

\paragraph{Bridge sampling.}
Each bridge sampling region aims to connect two seeds and thereby improve inter-region connectivity. Two seeds $s_i$ and $s_j$ form a \textit{visible pair} if the straight-line interpolation between them lies entirely in $\mathcal{Q}_\text{free}$. For each visible pair, we sample points along this line, compute at each point a noise radius that keeps the local collision-sample ratio within a predefined range, and draw bridge samples $\mathbf{q}_\text{bridge}$ within these radii, yielding $\mathcal{L}_\text{bridge} = \mathrm{BCE}(C_\text{gt},\, C(g_\theta(\mathbf{q}_\text{bridge})))$.

\subsection{Test-time Refinement}
\label{sec:refinement}

Learning from finite samples can leave safety-critical false positives in the predicted region, which we correct at test time by adjusting only the responsible explicit boundary parameters, without retraining.
When a false positive configuration $\mathbf{q}_\text{fp}$ is encountered during deployment, that is, a configuration predicted as collision-free that is actually in collision, the responsible boundary can be tightened by shifting its bias parameter inward:
\begin{equation}
    \hat{d}_{k,i} \leftarrow d_{k,i} - \Delta_{k,i},
\end{equation}
where $\Delta_{k,i}$ is computed from the signed distances of accumulated false positive points to hyperplane $i$ of region $k$.
For each false positive $\mathbf{z}_\text{fp}$, only the boundary with the smallest positive signed distance to it is held responsible. Let $\mathcal{F}_{k,i}$ collect the false positives so assigned to boundary $i$ of region $k$. Each boundary is then pulled inward by the worst violation across its assigned set, $\Delta_{k,i} = \max_{\mathbf{z}_\text{fp}\,\in\,\mathcal{F}_{k,i}} \phi_{k,i}(\mathbf{z}_\text{fp})$.
This operation shrinks only the responsible boundary, leaving all other regions unchanged. This mechanism is aligned with the train-time BCE objective under mild assumptions (Appendix~\ref{app:bce_gradient}).

\section{Results}
\label{sec:result}

\begin{figure}[t]
    \centering
    \newlength{\envFigH}\setlength{\envFigH}{0.24\linewidth}
    \begin{subfigure}[b]{0.247\linewidth}\centering
        \includegraphics[height=\envFigH,trim={0 0 0 25},clip]{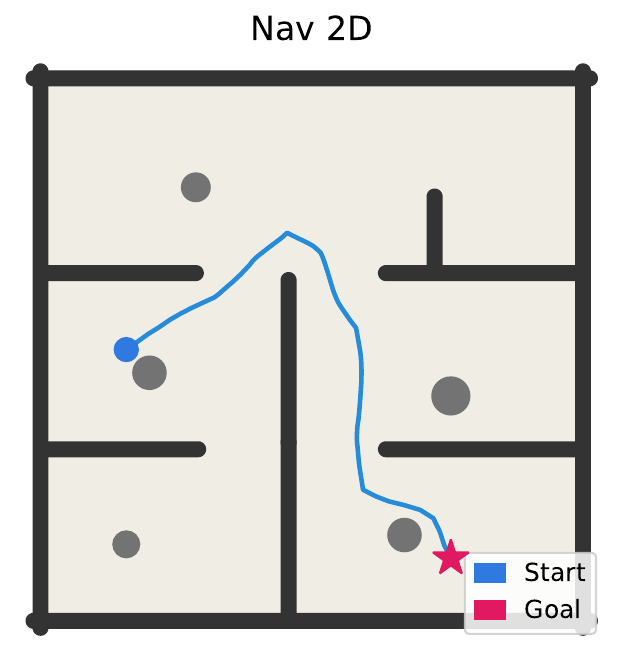}
        \caption{Nav 2D}
    \end{subfigure}\hfill
    \begin{subfigure}[b]{0.247\linewidth}\centering
        \includegraphics[height=\envFigH,trim={0 0 0 25},clip]{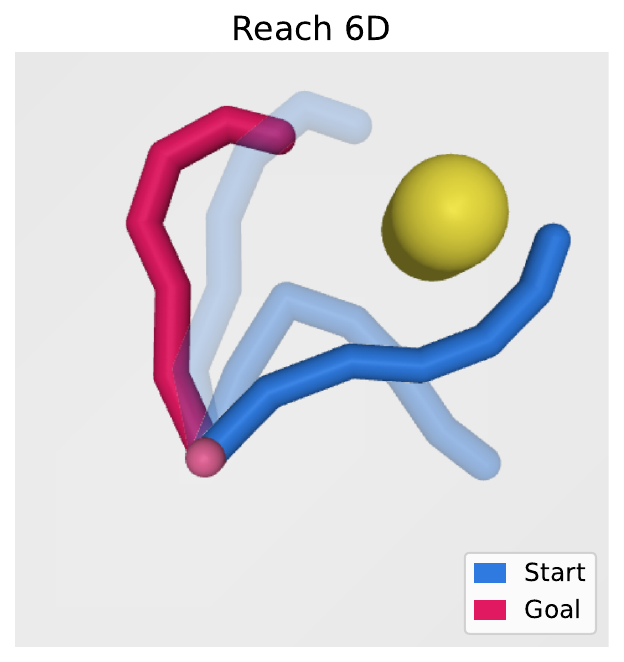}
        \caption{Reach 6D}
    \end{subfigure}\hfill
    \begin{subfigure}[b]{0.247\linewidth}\centering
        \includegraphics[height=\envFigH,trim={0 0 0 25},clip]{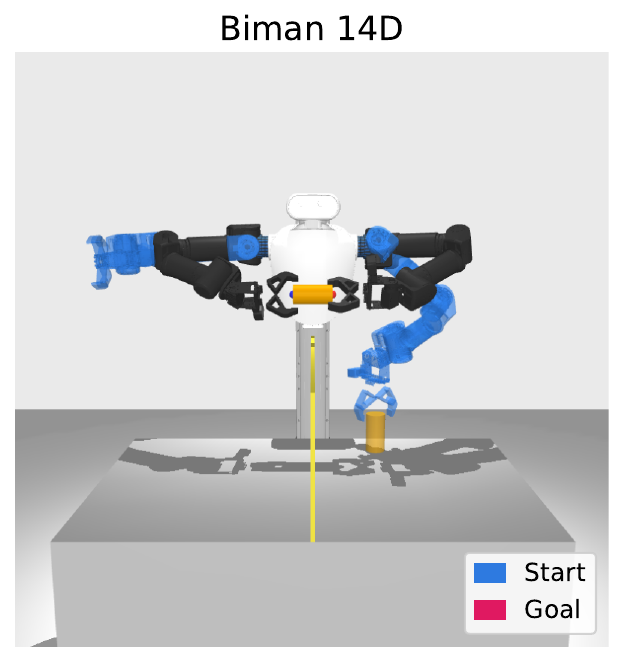}
        \caption{Bimanual 14D}
    \end{subfigure}\hfill
    \begin{subfigure}[b]{0.247\linewidth}\centering
        \includegraphics[height=\envFigH]{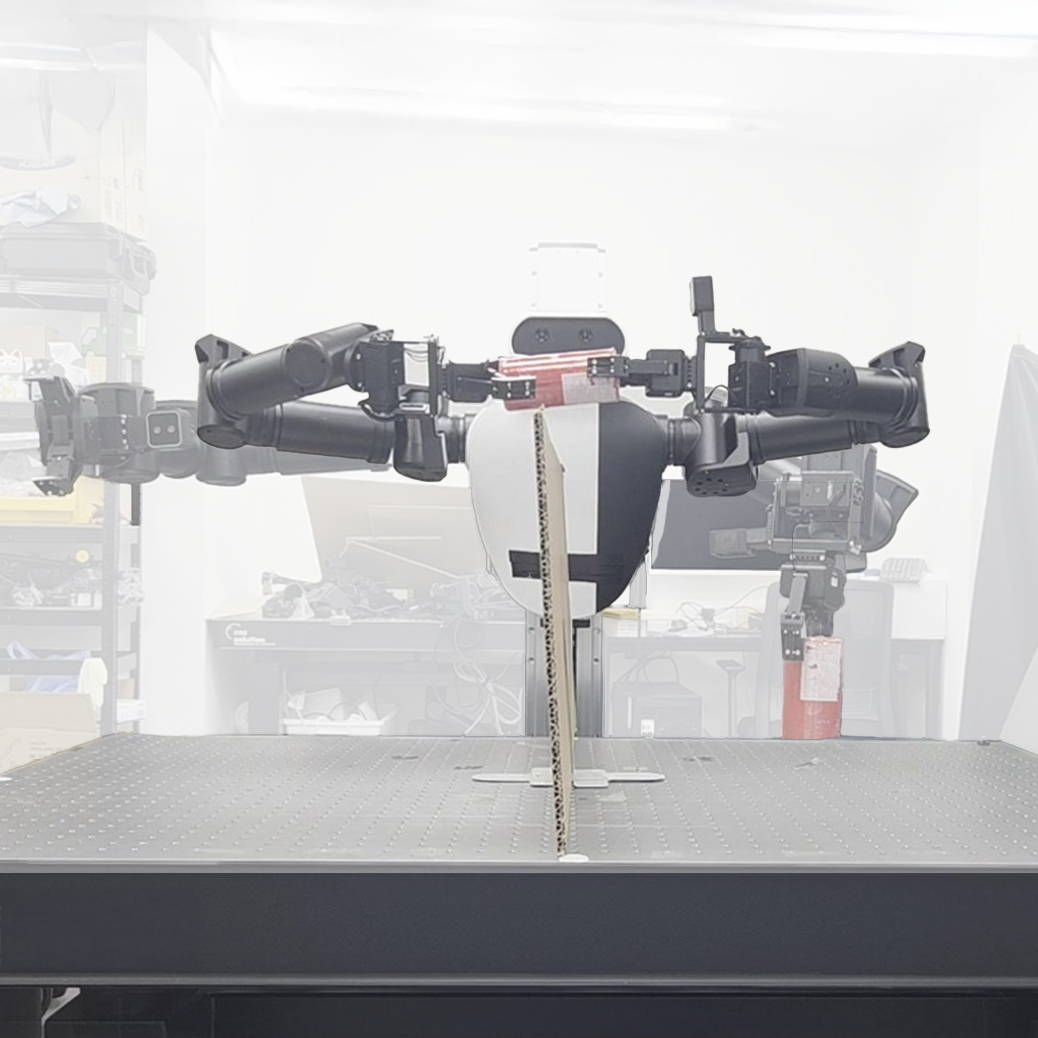}
        \caption{Bimanual 14D (Real)}
    \end{subfigure}
    \caption{
        Evaluation environments.
        Nav~2D is a 2D maze navigation task with circular obstacles and wall segments.
        Reach~6D is a 6-DoF planar reacher task requiring collision-free motion around a cylindrical obstacle.
        Reach~14D extends Reach~6D to 14-DoF and is omitted from the figure for space.
        Bimanual~14D is evaluated in both simulation and on a real-world 14-DoF bimanual manipulator.
    }
    \label{fig:env}
    \vspace{-8pt}
\end{figure}

We evaluate the proposed method from three complementary perspectives.
First, we show that the proposed representation achieves broader coverage and better inter-set connectivity than baselines, while retaining zero observed false positives after test-time refinement.
Second, we demonstrate that the proposed method is suitable for downstream planning, resulting in improved path-planning success rates.
Finally, we show that the proposed framework naturally supports test-time refinement, enabling real-time collision-free planning and practical adaptation to changes in scene geometry on a real bimanual manipulator setup.

\noindent\textbf{Environments.}
As illustrated in \Cref{fig:env}, we consider four environments: Nav~2D, Reach~6D, Reach~14D, and Bimanual~14D.
In all environments, the goal is to represent the collision-free region and plan a collision-free path from a collision-free start configuration $\mathbf{q}_\text{s}$ to a goal configuration $\mathbf{q}_\text{g}$.

\noindent\textbf{Baselines.}
For the representation evaluation (Section~\ref{sec:result_repr}), we compare our framework against IRIS~\cite{werner2024faster}, a standard geometric convex-decomposition baseline, and IRIS-Sel, a variant matched to our region budget.
We also conduct an ablation study using four methods, IRIS-Sel\,$\to$\,Q-Cvx\,$\to$\,Vis-Cvx\,$\to$\,ILD, which form a progressive chain where each iteration introduces a single core component on top of the previous baseline.
Specifically, Q-Cvx~\cite{deng2020cvxnet} replaces geometric construction with a learning-based decomposition that fits convex sets directly within the original configuration space.
Vis-Cvx introduces Visibility-Guided Sampling (VGS) to ensure topological inter-region connectivity.
Finally, ILD (ours) incorporates the invertible latent mapping on top of these features.

For the path-planning evaluation (Section~\ref{sec:result_planning}), every region-based method above is paired with a Graph of Convex Sets (GCS) planner~\cite{marcucci2023motion, drake}, which routes a path through a union of convex sets.
We additionally compare against widely used path planners that operate directly on the configuration space without an intermediate representation.
RRT-Connect~\cite{kuffner2000rrtconnect} and PRM~\cite{kavraki1996prm} are classical sampling-based planners.
cuRobo~\cite{sundaralingam2023curobo} is a GPU-accelerated path planner that we evaluate in two modes, its default mode (cuRobo) and its trajectory-optimization mode (cuRobo-Traj).

\noindent\textbf{Metrics.}
For the representation evaluation (Section~\ref{sec:result_repr}), we use three coverage metrics measuring the fraction of ground-truth collision-free configurations covered by the predicted region.
$\text{Cvg}_q$ is sample-based, counting the fraction of sampled collision-free configurations covered.
$\text{Cvg}_p$ and $\text{Cvg}_r$ are voxel-based, voxelizing the end-effector workspace (position and full pose, respectively) and counting the fraction of voxels reachable by collision-free configurations that are also reachable by the predicted region.
All three count only configurations in the \emph{largest} connected component, since planning is feasible only within one.
Nav~2D is a point robot, so we omit $\text{Cvg}_p$ and $\text{Cvg}_r$ for that environment.
We also report $N_\text{Isl}$ and $N$, the number of connected components and convex sets. For example, $N_\text{Isl}{=}1$ with $N{=}10$ means all ten sets are connected.
Details are in Appendix~\ref{app:coverage_metric}.
For the path-planning evaluation (Section~\ref{sec:result_planning}), we report path-planning success rate (SR) and solve time.

\noindent\textbf{Training details.}
We train $g_\theta$ (24 invertible blocks, hidden dimension $32$) jointly with the boundary parameters using Adam~\citep{kingma2015adam} for 10 epochs of 1000 iterations with batch size 1024.
Outside (collision) samples are weighted $10\times$ in the BCE loss to enforce zero false positives, and per-environment hyperparameters are listed in Appendix~\ref{app:hyperparameters}.

\subsection{Evaluating Collision-free Region Coverage and Connectivity} \label{sec:result_repr}

\begin{table*}[!h]
  \vspace{-8pt}
  \centering
\caption{
Collision-free region quality, pre- and post-refinement.
Top group (\emph{pre-refinement}) reports each method's raw decomposition;
bottom group applies test-time refinement (Section~\ref{sec:refinement}) to enforce precision\,=\,1.0.
We report precision (Prec.), coverage ($\text{Cvg}_q$, $\text{Cvg}_p$, $\text{Cvg}_r$), the number of convex sets ($N$), and connectivity as the number of connected components ($N_\text{Isl}$).
Among the refined methods, best $\text{Cvg}_q$ / $\text{Cvg}_p$ / $\text{Cvg}_r$ and smallest $N_\text{Isl}$ (ties included) are bolded.
}
\label{tab:repr_flat}

  \scriptsize
  \setlength{\tabcolsep}{2.0pt}
  \renewcommand{\arraystretch}{0.95}

  \resizebox{\textwidth}{!}{
  \begin{tabular}{l cccr cccccr cccccr cccccr}
  \toprule
  & \multicolumn{4}{c}{Nav~2D}
  & \multicolumn{6}{c}{Reach~6D}
  & \multicolumn{6}{c}{Reach~14D}
  & \multicolumn{6}{c}{Bimanual~14D} \\
  \cmidrule(lr){2-5} \cmidrule(lr){6-11} \cmidrule(lr){12-17} \cmidrule(lr){18-23}
  Method
  & Prec. & $\text{Cvg}_q$ & $N_\text{Isl}$ & $N$
  & Prec. & $\text{Cvg}_q$ & $\text{Cvg}_p$ & $\text{Cvg}_r$ & $N_\text{Isl}$ & $N$
  & Prec. & $\text{Cvg}_q$ & $\text{Cvg}_p$ & $\text{Cvg}_r$ & $N_\text{Isl}$ & $N$
  & Prec. & $\text{Cvg}_q$ & $\text{Cvg}_p$ & $\text{Cvg}_r$ & $N_\text{Isl}$ & $N$ \\
  \midrule
  \multicolumn{23}{l}{\textit{Pre-refinement}} \\
  IRIS~\cite{werner2024faster}
    &  99.98 &  89.7 &  1 & 29
    &  92.99 &  88.8 & 99.4 & 93.8 &  1 & 52
    &  76.37 &  82.2 & 99.8 & 99.4 &  2 & 27
    &  63.41 &  77.1 & 77.1 & 77.1 &  2 & 90 \\
  IRIS-Sel
    &  99.99 &  89.2 &  1 & 18
    &  97.39 &  61.2 & 65.8 & 57.6 &  2 &  3
    &  84.20 &  54.1 & 99.7 & 97.0 &  1 &  3
    &  74.47 &  36.2 & 36.3 & 36.2 &  1 &  3 \\
  Q-Cvx~\cite{deng2020cvxnet}
    &  94.57 &  65.6 & 11 & 18
    &  99.98 &  53.3 & 61.8 & 50.3 &  2 &  3
    & 100.00 &   2.7 & 45.5 & 26.4 &  3 &  3
    &  99.76 &  22.6 & 56.7 & 25.7 &  3 &  3 \\
  Vis-Cvx
    &  66.22 &  88.3 &  2 & 18
    &  99.90 &  39.2 & 60.7 & 46.4 &  2 &  3
    &  99.44 &  19.7 & 96.5 & 84.7 &  1 &  3
    &  95.18 &  60.9 & 88.0 & 63.8 &  1 &  3 \\
  ILD (ours)
    &  96.75 &  96.1 &  1 & 18
    &  99.13 &  87.0 & 98.7 & 90.8 &  1 &  3
    &  99.65 &  37.1 & 98.0 & 93.0 &  1 &  3
    &  99.50 &  43.7 & 74.6 & 46.7 &  1 &  3 \\
  \midrule
  \multicolumn{23}{l}{\textit{Post-refinement (precision\,=\,1.0)}} \\
  IRIS~\cite{werner2024faster}
    & 100.00 &  89.2 & \textbf{1} & 29
    & 100.00 &  25.6 & 85.6 & 54.9 & \textbf{1} & 52
    & 100.00 &   0.2 & 33.9 &  7.1 &  2 & 27
    & 100.00 &   0.2 &  2.5 &  0.3 &  6 & 90 \\
  IRIS-Sel
    & 100.00 &  88.6 & \textbf{1} & 18
    & 100.00 &  15.3 & 49.2 & 30.0 &  2 &  3
    & 100.00 &   0.1 & 17.2 &  3.4 & \textbf{1} &  3
    & 100.00 &   0.1 &  0.7 &  0.1 & \textbf{1} &  3 \\
  Q-Cvx~\cite{deng2020cvxnet}
    & 100.00 &  58.0 & 11 & 18
    & 100.00 &  52.9 & 61.7 & 49.9 &  2 &  3
    & 100.00 &   2.7 & 45.5 & 26.4 &  3 &  3
    & 100.00 &  11.1 & 38.9 & 13.5 &  3 &  3 \\
  Vis-Cvx
    & 100.00 &  48.3 &  3 & 18
    & 100.00 &  36.7 & 57.8 & 43.9 &  2 &  3
    & 100.00 &  10.2 & 79.1 & 58.4 &  2 &  3
    & 100.00 &  14.8 & 46.4 & 17.6 & \textbf{1} &  3 \\
  ILD (ours)
    & 100.00 & \textbf{91.9} & \textbf{1} & 18
    & 100.00 & \textbf{75.5} & \textbf{96.0} & \textbf{81.8} & \textbf{1} &  3
    & 100.00 & \textbf{22.2} & \textbf{92.7} & \textbf{80.0} & \textbf{1} &  3
    & 100.00 & \textbf{22.1} & \textbf{51.7} & \textbf{24.9} & \textbf{1} &  3 \\
  \bottomrule
  \end{tabular}
  }
  \end{table*}

We evaluate the quality of the collision-free region representation for each method on a uniformly sampled set of configurations from $\mathcal{Q}_\text{free}$.
\Cref{tab:repr_flat} reports both the raw decomposition and its refined counterpart after applying test-time refinement (Section~\ref{sec:refinement}) to observed false positives on this evaluated set, followed by three analyses along the baseline chain.

\noindent\textbf{\textit{Learning-based decomposition improves coverage in high dimensions}} (IRIS-Sel vs.\ Q-Cvx).
At a matched region count, the learning-based Q-Cvx more than quadruples the mean $\text{Cvg}_q$ of the geometric IRIS-Sel on the three manipulation environments ($22.2$ vs.\ $5.2$), and the gap widens according to the configuration-space dimension.
Only in low-dimensional Nav~2D does the geometric baseline remain competitive, indicating that learning-based decomposition scales much better with dimensionality.

\noindent\textbf{\textit{VGS improves connectivity}} (Q-Cvx vs.\ Vis-Cvx).
Adopting VGS seeds (Vis-Cvx) in place of Q-Cvx's random seeds more than halves the mean number of islands ($N_\text{Isl}\!:4.75 \!\to\! 2.0$), with the largest drop on Nav~2D ($11\!\to\!3$).
VGS therefore improves the connectivity of the decomposition.

\noindent\textbf{\textit{Invertible latent mapping further improves coverage}} (Vis-Cvx vs.\ ILD).
Adding the invertible latent mapping on top of Vis-Cvx yields ILD's largest gain. It nearly doubles $\text{Cvg}_q$ averaged across the four environments ($27.5 \!\to\! 52.9$), and the same ranking holds in task space ($\text{Cvg}_p$, $\text{Cvg}_r$).
The latent deformation is therefore what lets a compact set of convex regions conform to highly non-convex free regions in $\mathcal{Q}$.

\subsection{Evaluating Downstream Path Planning}
\label{sec:result_planning}

\begin{table*}[!h]
\vspace{-8pt}
\centering
\caption{
Motion planning evaluation with and without test-time refinement.
For every environment, we report two rows. The top row, labeled \emph{pre-refinement}, gives the first-evaluation pass \textit{before} test-time refinement, and the bottom row gives the result \textit{after} refinement. 
Sampling-based baselines do not refine and are compared against the post-refinement rows.
Higher SR and lower Time are better. Bold marks the best post-refinement result per row.
}
\label{tab:planning_eval}
\scriptsize
\setlength{\tabcolsep}{2.5pt}
\renewcommand{\arraystretch}{0.95}
\resizebox{\textwidth}{!}{
\begin{tabular}{ll cccc ccccc}
\toprule
 & & \multicolumn{4}{c}{Sampling-based} & \multicolumn{5}{c}{Region-based} \\
\cmidrule(lr){3-6} \cmidrule(lr){7-11}
Env. & Metric
& RRT~\cite{kuffner2000rrtconnect}
& PRM~\cite{kavraki1996prm}
& cuRobo~\cite{sundaralingam2023curobo}
& cuRobo-Traj~\cite{sundaralingam2023curobo}
& IRIS~\cite{werner2024faster}
& IRIS-Sel
& Q-Cvx~\cite{deng2020cvxnet}
& Vis-Cvx
& ILD (ours) \\
\midrule

\multirow{2}{*}{\makecell[l]{Nav 2D\\\scriptsize(pre-refinement)}}
& SR (\%)  & - & - & - & - & 94.0 & 98.2 & 25.1 & 14.3 & 98.3 \\
& Time (s) & - & - & - & - & 0.17 & 0.07 & 0.07 & 0.09 & 0.09 \\
\multirow{2}{*}{Nav 2D}
& SR (\%)  & 98.5 & 98.3 & 93.0 & 13.2 & \textbf{100.0} & 98.2 & 31.8 & 72.8 & 99.7 \\
& Time (s) & 0.002 & \textbf{0.001} & 0.19 & 4.25 & 0.06 & 0.03 & 0.02 & 0.05 & 0.06 \\
\midrule

\multirow{2}{*}{\makecell[l]{Reach 6D\\\scriptsize(pre-refinement)}}
& SR (\%)  & - & - & - & - & 92.4 & 4.0 & 98.1 & 68.5 & 98.9 \\
& Time (s) & - & - & - & - & 18.58 & 0.02 & 0.02 & 0.03 & 0.20 \\
\multirow{2}{*}{Reach 6D}
& SR (\%)  & 99.1 & 99.1 & 97.2 & 3.0 & \textbf{99.9} & 4.0 & 98.1 & 99.5 & \textbf{99.9} \\
& Time (s) & \textbf{0.003} & 0.004 & 0.10 & 4.61 & 21.17 & 0.02 & 0.02 & 0.03 & 0.16 \\
\midrule

\multirow{2}{*}{\makecell[l]{Reach 14D\\\scriptsize(pre-refinement)}}
& SR (\%)  & - & - & - & - & 2.4 & 1.3 & 28.7 & 91.9 & 97.7 \\
& Time (s) & - & - & - & - & 47.93 & 1.10 & 2.44 & 1.27 & 0.88 \\
\multirow{2}{*}{Reach 14D}
& SR (\%)  & 97.0 & 94.9 & 90.1 & 14.5 & 2.4 & 1.3 & 28.7 & 92.0 & \textbf{97.7} \\
& Time (s) & \textbf{0.004} & 0.010 & 0.10 & 4.99 & 47.93 & 1.10 & 0.35 & 1.16 & 0.88 \\
\midrule

\multirow{2}{*}{\makecell[l]{Bimanual~14D\\\scriptsize(pre-refinement)}}
& SR (\%)  & - & - & - & - & 92.0 & 36.7 & 51.0 & 35.9 & 91.2 \\
& Time (s) & - & - & - & - & 365.46 & 1.12 & 0.04 & 0.15 & 0.36 \\
\multirow{2}{*}{Bimanual~14D}
& SR (\%)  & 88.6 & 88.4 & 73.5 & 46.0 & 96.0 & 36.7 & 82.3 & 97.1 & \textbf{98.4} \\
& Time (s) & \textbf{0.002} & 0.022 & 0.06 & 2.49 & 75.01 & 1.12 & 0.05 & 0.19 & 0.35 \\
\bottomrule
\end{tabular}
}
\vspace{-4pt}
\end{table*}

We evaluate whether the proposed representation enables effective downstream path planning.
Each baseline uses its standard collision representation, with the collision margin tuned per environment (Appendix~\ref{app:baseline_tuning}).
\Cref{tab:planning_eval} reports the results across all environments.

\paragraph{Success rate.}
ILD achieves the highest SR across all environments, and the chain Q-Cvx\,$\to$\,Vis-Cvx\,$\to$\,ILD shows that both VGS and the invertible mapping contribute.
IRIS drops in high dimensions because its post-refinement coverage is too low.
RRT, PRM, and cuRobo rely on primitive (sphere/capsule) obstacle proxies, which inflate obstacle volume and close off tight passages, so their SR drops in Reach~14D and Bimanual~14D, whereas ILD stays near $98\%$ by routing on the learned safe region that absorbs the full mesh geometry.
Region-based methods apply test-time refinement, which raises SR consistently across all environments.

\paragraph{Solve time.}
ILD outpaces IRIS by a widening margin as dimensionality grows. Averaged across the two 14D experiments (Reach~14D and Bimanual~14D), ILD solves in $0.45$\,s versus IRIS's $51.83$\,s, a $\sim 115\times$ speedup.
The gap stems from ILD's fewer, better-connected regions, which shrink the GCS graph the planner has to search.
ILD is slower in absolute terms than the sampling-based baselines, but that gap reflects the cost those baselines avoid by approximating obstacles as primitives. ILD does not rely on such an approximation, which is consistent with its higher SR.

\subsection{Evaluating Test-time Refinement in the Real World}
\label{sec:result_realworld}

\begin{figure}[h]
    \centering
    \begin{minipage}[c]{0.46\linewidth}\centering
        \begin{minipage}{0.49\linewidth}\centering
            \includegraphics[width=\linewidth]{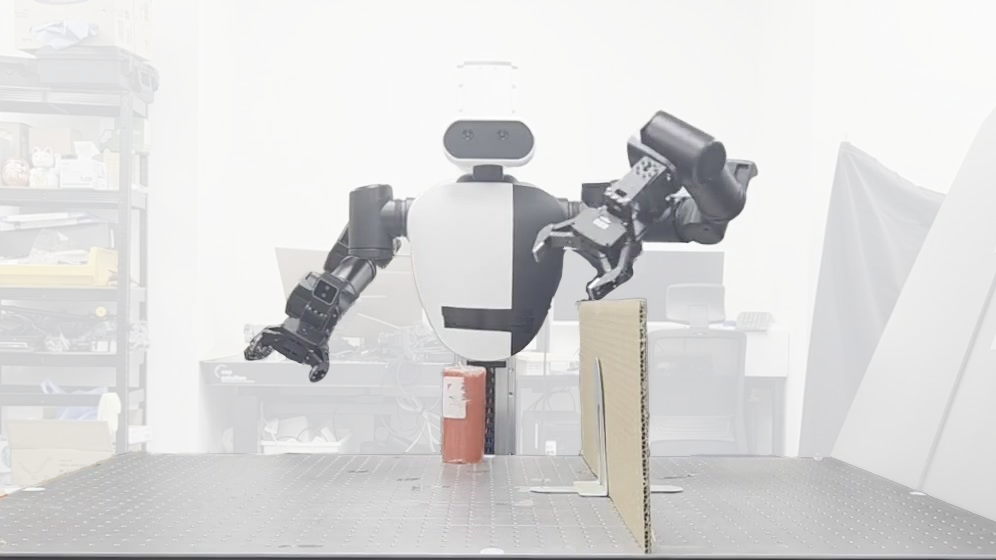}\\
            \scriptsize approach
        \end{minipage}\hfill
        \begin{minipage}{0.49\linewidth}\centering
            \includegraphics[width=\linewidth]{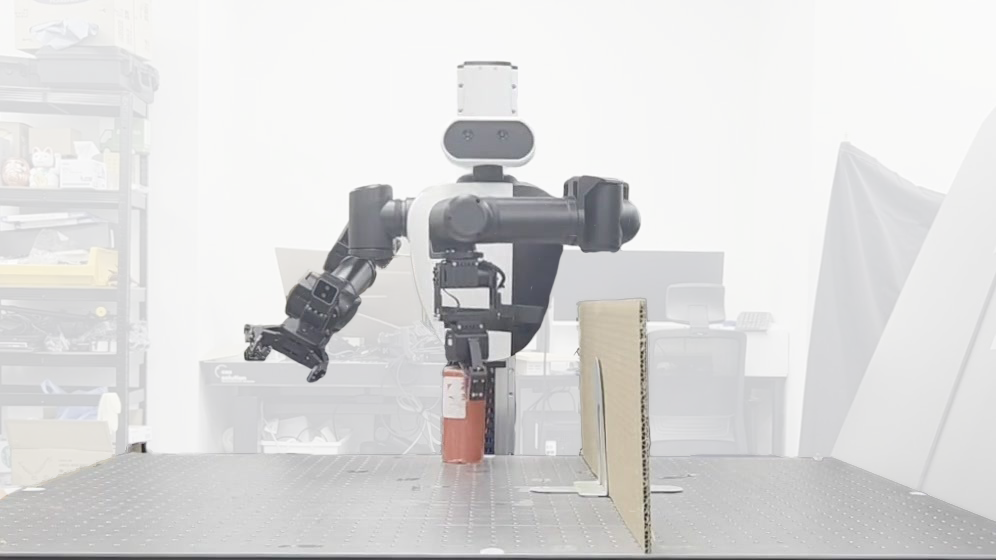}\\
            \scriptsize grasp
        \end{minipage}\\[2pt]
        \small Training env.\ ($h{=}0.20$\,m)
    \end{minipage}
    \hfill\rule[-0.09\linewidth]{0.4pt}{0.18\linewidth}\hfill
    \begin{minipage}[c]{0.46\linewidth}\centering
        \begin{minipage}{0.49\linewidth}\centering
            \includegraphics[width=\linewidth]{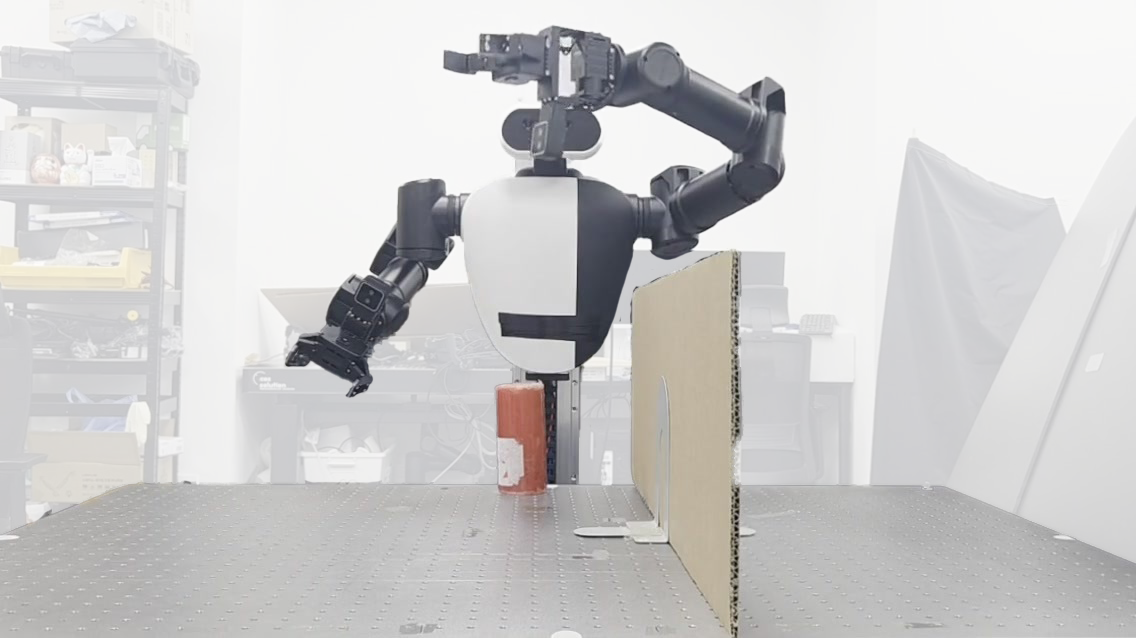}\\
            \scriptsize approach
        \end{minipage}\hfill
        \begin{minipage}{0.49\linewidth}\centering
            \includegraphics[width=\linewidth]{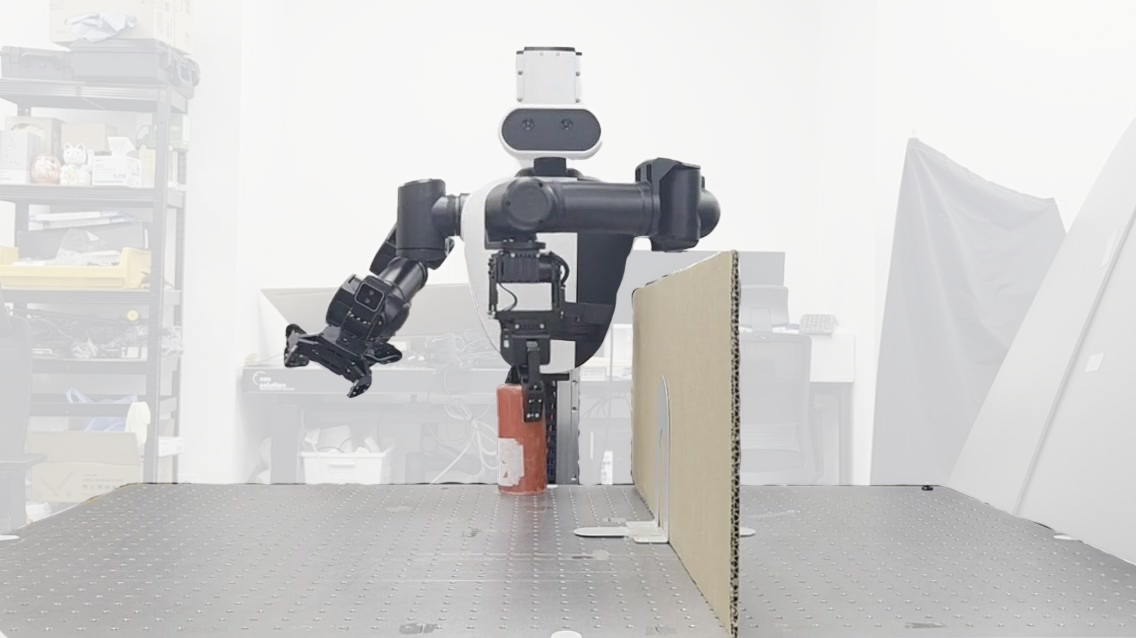}\\
            \scriptsize grasp
        \end{minipage}\\[2pt]
        \small Test-time env.\ ($h{=}0.30$\,m)
    \end{minipage}
    \caption{Test-time refinement under wall height changes. ILD is first
    trained with the wall at $h{=}0.20$\,m and then refined at deployment
    to $h{=}0.30$\,m without retraining. As the wall is raised, the refined
    boundary forces the arm to arc higher while reaching the same grasp.}
    \label{fig:realworld_refine}
    \vspace{-8pt}
\end{figure}

We demonstrate that test-time refinement enables ILD to adapt to changes in scene geometry on a real bimanual manipulator (\Cref{fig:realworld_refine}). The task is to pick an object from a tabletop while clearing a wall between the base and the target. ILD is first trained with the wall at $h{=}0.20$\,m and then evaluated on the same setup with the wall raised to $h{=}0.30$\,m without retraining. Test-time refinement (Section~\ref{sec:refinement}) is the only update applied at deployment.
At $h{=}0.30$\,m the arm arcs noticeably higher to clear the raised wall before settling into the same grasp, confirming that ILD's refined boundary transfers to deployment without retraining or hand-tuned waypoints.
As a side note, only a 6-DoF arm (one arm, wrist roll locked) of the bimanual manipulator is used in this deployment, unlike \Cref{fig:env}(d).

\section{Limitations}
\label{sec:limitations}

ILD has two main limitations. First, the convex decomposition is fit to a single static scene, so it does not handle scenes that change over time. Test-time refinement only contracts boundaries to absorb false positives observed in the current scene, which trades coverage for safety. A natural extension is to amortize this per-scene fit into a feedforward network that emits the boundary parameters from a scene representation, analogous to recent feedforward formulations of per-scene 3D representations~\citep{charatan2024pixelsplat, chen2024mvsplat}. Second, the latent space is invertible but not isometric, so a path that minimizes length in the latent space is not necessarily shortest in configuration space. The isometric regularizer encourages nearby configurations to remain nearby in the latent space and reduces this gap, but does not fully close it. Jointly training the planner and the latent mapping, so that the planning objective is defined directly in configuration space, is a promising direction.

\section{Conclusion}
\label{sec:conclusion}

Existing collision-free representations fall into two families with complementary strengths. Explicit representations expose hard collision-free constraints while implicit representations scale gracefully, but neither delivers both at once. We show that ILD (Invertible Latent Decomposition) bridges the two by combining an implicit invertible neural mapping with an explicit convex decomposition in the resulting latent space. Our framework retains explicit collision-free constraints and supports real-time deployment that adapts to scene-geometry changes. Our results show that ILD attains the highest path-planning success rate across all environments while solving over $100\times$ faster than the convex-decomposition baseline~\cite{werner2024faster} at 14-DoF. Together, these results suggest that an invertible latent decomposition bridges the explicit and implicit families into an effective collision-free space representation for path planning.

\clearpage
\acknowledgments{If a paper is accepted, the final camera-ready version will (and probably should) include acknowledgments. All acknowledgments go at the end of the paper, including thanks to reviewers who gave useful comments, to colleagues who contributed to the ideas, and to funding agencies and corporate sponsors that provided financial support.}

\bibliography{root}

\clearpage
\appendix
\section*{Appendix Overview}

This appendix expands on the technical details, deployment procedure, and experimental setup of ILD.
\Cref{app:training} covers training details, including the architecture of the invertible mapping, Visibility-Guided Sampling (VGS), regularization losses, and training hyperparameters.
\Cref{app:path_planning} covers deployment details, including the test-time refinement procedure, start- and goal-snap bridging, and redundant facet pruning.
\Cref{app:details} covers additional experimental details, including the coverage metric definitions, planning evaluation setup, path-planning dataset, full representation quality table, and baseline hyperparameter tuning.
\Cref{app:bce_gradient} provides a geometric interpretation of the joint objective, including its connection to test-time refinement.

\section{Training Details}
\label{app:training}

\subsection{Invertible Latent Mapping: Architecture and Initialization}
\label{app:architecture}

\noindent\textbf{Architecture.}
We model the invertible mapping $g_\theta : \mathcal{Q} \to \mathcal{Z}$ using an architecture similar to those used in normalizing flows~\citep{rezende2015variational, kobyzev2020normalizing}, composed of alternating invertible linear layers and affine coupling layers~\citep{kingma2018glow}.
Concretely, the network has the following structure:
\begin{equation*}
    g_\theta = \underbrace{W_0}_{\text{LU}} \to \left[\underbrace{\mathrm{Coupling}_1 \to W_1}_{\text{block}}\right] \to \cdots \to \left[\underbrace{\mathrm{Coupling}_L \to W_L}_{\text{block}}\right],
\end{equation*}
where $L$ is the number of blocks.
We use $L = 24$ blocks for planning experiments, giving a total of 49 alternating layers.

\noindent\textbf{Invertible linear layer (LU decomposition).}
Each $W_i$ is an invertible linear layer parameterized via LU decomposition:
\begin{equation}
    W = P L U,
\end{equation}
where $P \in \mathbb{R}^{n \times n}$ is a fixed permutation matrix, $L$ is a lower-triangular matrix with unit diagonal, and $U$ is an upper-triangular matrix with learnable diagonal entries $S = \mathrm{diag}(U)$ stored in log-scale for numerical stability.
The forward pass is $\mathbf{z} = \mathbf{x} W$ and the inverse is $\mathbf{x} = \mathbf{z} W^{-1}$, with $\log |\det W| = \sum_i \log |S_i|$.

\noindent\textbf{Affine coupling layer.}
Each coupling layer splits the input $\mathbf{x} \in \mathbb{R}^n$ into two halves $(\mathbf{x}_1, \mathbf{x}_2)$ and applies:
\begin{equation}
    \mathbf{y}_1 = \mathbf{x}_1, \qquad
    \mathbf{y}_2 = \mathbf{x}_2 \odot \exp(s(\mathbf{x}_1)) + t(\mathbf{x}_1),
\end{equation}
where $s(\cdot)$ (scale) and $t(\cdot)$ (translate) are small MLPs of the form Linear($n/2$, $h$) $\to$ ReLU $\to$ Linear($h$, $n/2$), with hidden dimension $h = 32$.
The scale network additionally applies Tanh to bound the scaling factor.
The inverse is $\mathbf{x}_2 = (\mathbf{y}_2 - t(\mathbf{y}_1)) \odot \exp(-s(\mathbf{y}_1))$.

\noindent\textbf{Initialization: isometric initialization.}
Both layer types are initialized so that $g_\theta$ begins as an orthogonal (distance-preserving) map.
For each LU layer, a random matrix is QR-decomposed to obtain an orthogonal matrix $Q$, whose LU decomposition directly initializes $W = Q$.
For each affine coupling layer, the final linear layers of $s(\cdot)$ and $t(\cdot)$ are zero-initialized, so $s = \mathbf{0}$ and $t = \mathbf{0}$ at the start, making each coupling layer the identity.
In addition, $g_\theta$ starts as a composition of orthogonal maps, ensuring $\|g_\theta(\mathbf{x}) - g_\theta(\mathbf{y})\| = \|\mathbf{x} - \mathbf{y}\|$ exactly at initialization.
This isometric start is also what makes the isometric loss (Section~\ref{app:reg_losses}) meaningful. By anchoring $g_\theta$ to the initial snapshot $g_{\theta_0}$ and penalizing pairwise distance distortion, the mapping remains approximately isometric throughout training.

\subsection{Visibility-Guided Sampling (VGS)}
\label{app:seed_selection}

VGS has two stages, an offline precompute that runs once before training and a draw-only loop that runs every iteration. The offline stage selects seed configurations in $\mathcal{Q}_\text{free}$ (Section~\ref{sec:vgs}) and prepares the training samples used by the seed and bridge losses. Every training iteration then draws mini-batches directly from this precomputed cache without invoking the collision oracle, so the per-iteration sampling cost stays negligible.

\paragraph{Seed selection.}
Algorithm~\ref{alg:seed_selection} gives the full pseudocode used in Section~\ref{sec:vgs}. We first sample a pool of candidate configurations uniformly from $\mathcal{Q}_\text{free}$ and build a symmetric visibility matrix $A$ by discretising each straight-line segment between candidates into $N_\text{intp} = 100$ interpolation points and querying the collision oracle on all of them in parallel. For each candidate we also compute its clearance to the nearest obstacle and convert it to a per-candidate rank in $[0, 1]$, which makes the interior bias scale-invariant across very different workspaces. The greedy loop then grows a seed set one element at a time by selecting the candidate with the largest score. For candidate $i$, let $\text{gain}_i$ denote the number of currently uncovered candidates that are visible from $i$ in $A$, i.e., the coverage that would be added if $i$ were chosen next. The score is then
\begin{equation*}
\text{score}_i \;=\; \text{gain}_i \cdot (1 + \alpha \cdot \text{interior\_rank}_i),
\end{equation*} 
The first seed may be any candidate, but every subsequent seed must itself be already covered (i.e., visible from a previously chosen seed), so the chosen seed set stays inside a single connected component of the visibility graph. The loop terminates when the cumulative recall exceeds a target threshold, the best remaining coverage gain is zero, or the seed budget is exhausted. In our experiments, we pick seeds until the cumulative ratio of covered candidates reaches a target ratio $\rho_\text{trgt}$, set to $0.99$ for Nav~2D and $0.95$ for the manipulation environments (Reach~6D, Reach~14D, Bimanual~14D).

\begin{algorithm}[h]
\caption{Greedy visibility-based seed selection.\\
\textbf{Inputs:} candidate pool of size $n$ with visibility matrix $A$; per-candidate interior-rank in $[0,1]$; seed budget $N_\text{seed}$; target recall $\rho_\text{trgt}$; interior weight $\alpha$.\\
\textbf{Output:} seed list $\mathbf{s}$.}
\label{alg:seed_selection}
\begin{algorithmic}[1]
\State $\mathbf{s} \gets \emptyset$, $\;\mathcal{V} \gets \emptyset$ \Comment{seed list and covered set}
\For{$\text{step} = 1, \dots, N_\text{seed}$}
    \State $E \gets \{1, \dots, n\}$ if $\mathbf{s} = \emptyset$ else $E \gets \mathcal{V}$ \Comment{eligible candidates}
    \For{each $i \in E$}
        \State $\text{gain}_i \gets |\{j \notin \mathcal{V} : A_{i,j} = 1\}|$
        \State $\text{score}_i \gets \text{gain}_i \cdot (1 + \alpha \cdot \text{interior\_rank}_i)$
    \EndFor
    \State $i^\star \gets \arg\max_{i \in E} \text{score}_i$
    \State $\mathbf{s} \gets \mathbf{s} \cup \{i^\star\}$, $\;\mathcal{V} \gets \mathcal{V} \cup \{j : A_{i^\star, j} = 1\}$
    \If{$|\mathcal{V}|/n \ge \rho_\text{trgt}$} \textbf{break} \EndIf
\EndFor
\State \Return $\mathbf{s}$
\end{algorithmic}
\end{algorithm}

\paragraph{Seed sampling.}
After seed selection, we sample for the per-seed BCE terms $\mathcal{L}_\text{seed}$ and $\mathcal{L}_\text{vis}$ in Section~\ref{sec:vgs} as described in Algorithm~\ref{alg:seed_sampling}. For each chosen seed $s_k$, $\mathbf{q}_\text{seed}$ is drawn from a fixed-radius neighborhood around $s_k$ with labels assigned by $C_\text{gt}(\cdot)$, while $\mathbf{q}_\text{vis}$ is drawn uniformly from the candidates marked visible from $s_k$ by the visibility matrix $A$ and labeled as collision-free by construction. The visible candidates are cached once after seed selection, so every training iteration draws both mini-batches without re-running any collision query.

\begin{algorithm}[H]
\caption{Seed sampling for one seed.\\
\textbf{Inputs:} seed $s_k$; cached visible candidates $\mathcal{V}_k$; neighborhood radius $r_\text{nb}$; sample counts $T_\text{seed}, T_\text{vis}$; collision check $C_\text{gt}$.\\
\textbf{Output:} (sample, label) pairs $\mathcal{D}$.}
\label{alg:seed_sampling}
\begin{algorithmic}[1]
\State $\mathcal{D} \gets \emptyset$
\For{$\tau = 1, \dots, T_\text{seed}$}
    \State $\boldsymbol{\epsilon} \gets$ sample from $\mathrm{Uniform}([-r_\text{nb}, r_\text{nb}]^n)$
    \State $\mathbf{q}_\tau \gets s_k + \boldsymbol{\epsilon}$
    \State $y_\tau \gets C_\text{gt}(\mathbf{q}_\tau)$
    \State $\mathcal{D} \gets \mathcal{D} \cup \{(\mathbf{q}_\tau, y_\tau)\}$
\EndFor
\For{$\tau = 1, \dots, T_\text{vis}$}
    \State $\mathbf{q}_\tau \gets$ uniform sample from $\mathcal{V}_k$
    \State $\mathcal{D} \gets \mathcal{D} \cup \{(\mathbf{q}_\tau, 1)\}$
\EndFor
\State \Return $\mathcal{D}$
\end{algorithmic}
\end{algorithm}

\paragraph{Bridge sampling.}
For each visible seed pair $(s_i, s_j)$ promoted to a bridge sampling region, we sample for the bridge BCE in Section~\ref{sec:vgs} as described in Algorithm~\ref{alg:bridge_sampling}. We discretise the straight-line segment between $s_i$ and $s_j$ into $M$ interpolation points $\{\mathbf{p}_m\}_{m=1}^{M}$. For each $\mathbf{p}_m$, we precompute a radius $r_m$ such that Gaussian noise centered at $\mathbf{p}_m$ with standard deviation $r_m$ keeps samples inside $\mathcal{Q}_\text{free}$ about 90\% of the time (so roughly 10\% land in collision). Each $\mathbf{q}_\text{bridge}$ is then drawn by picking $m$ uniformly at random from $\{1, \dots, M\}$ and adding Gaussian noise scaled by $r_m$, with the collision check $C_\text{gt}(\cdot)$ from Section~\ref{sec:lcd} assigning the binary label. The minority of in-collision samples gives the BCE a nearby negative signal, which prevents the learned region from drifting and producing false positives near the bridge.

\begin{algorithm}[H]
\caption{Bridge sampling for a visible pair.\\
\textbf{Inputs:} seed pair $(s_i, s_j)$; pre-computed interpolation points $\{\mathbf{p}_m\}_{m=1}^{M}$ and radii $\{r_m\}_{m=1}^{M}$; total sample count $T_\text{total}$; collision check $C_\text{gt}$.\\
\textbf{Output:} (sample, label) pairs $\mathcal{D}$.}
\label{alg:bridge_sampling}
\begin{algorithmic}[1]
\State $\mathcal{D} \gets \emptyset$
\For{$\tau = 1, \dots, T_\text{total}$}
    \State $m \gets$ uniform sample from $\{1, \dots, M\}$
    \State $\boldsymbol{\epsilon} \gets$ sample from $\mathcal{N}(\mathbf{0}, r_m^2 \mathbf{I})$
    \State $\mathbf{q}_\tau \gets \mathbf{p}_m + \boldsymbol{\epsilon}$
    \State $y_\tau \gets C_\text{gt}(\mathbf{q}_\tau)$
    \State $\mathcal{D} \gets \mathcal{D} \cup \{(\mathbf{q}_\tau, y_\tau)\}$
\EndFor
\State \Return $\mathcal{D}$
\end{algorithmic}
\end{algorithm}

\subsection{Regularization Losses}
\label{app:reg_losses}

Three additional regularization losses are applied during training to keep the latent mapping well-behaved and the region boundaries faithful to the collision-free criterion.

\noindent\textbf{Isometric loss.}
The isometric loss keeps the latent map $g_\theta$ close to a distance-preserving deformation of the configuration space.
Without it, the joint optimization of $(\theta, \mathcal{H})$ can let $g_\theta$ deform excessively, distorting the metric structure of $\mathcal{Z}$ and making paths that appear short in latent space correspond to unnecessarily long or winding trajectories once decoded back to $\mathcal{Q}$.
We use two terms, an anchor term that ties $g_\theta$ to its initial snapshot $g_{\theta_0}$ taken at the start of training,
\begin{equation}
    \mathcal{L}_\text{anc} \;=\; \mathbb{E}_{\mathbf{q}}\!\left[\,\| g_\theta(\mathbf{q}) - g_{\theta_0}(\mathbf{q}) \|^2\,\right],
\end{equation}
and a pairwise distance-preservation term over sampled pairs $(\mathbf{q}_1, \mathbf{q}_2)$,
\begin{equation}
    \mathcal{L}_\text{iso} \;=\; \mathbb{E}_{\mathbf{q}_1, \mathbf{q}_2}\!\left[\,\big(\| g_\theta(\mathbf{q}_1) - g_\theta(\mathbf{q}_2) \| - \| \mathbf{q}_1 - \mathbf{q}_2 \|\big)^2\,\right].
\end{equation}
Together, the two terms are introduced so that paths planned in $\mathcal{Z}$ stay near-optimal when decoded back to $\mathcal{Q}$.

\noindent\textbf{Box loss.}
The box loss enforces normalized joint-limit feasibility. We normalize the configuration space to $[-1, 1]^n$ and, for collision configurations $\mathbf{q}$ with $C_\text{gt} = 0$ that exceed the box of half-width $m \in (0, 1]$, weight the bound violation by the classifier output $C(g_\theta(\mathbf{q}))$:
\begin{equation}
    \mathcal{L}_\text{box} = \mathbb{E}_{\mathbf{q}\,:\,C_\text{gt}=0}\!\left[\, C(g_\theta(\mathbf{q})) \cdot \sum_i \mathrm{ReLU}\!\left(|\mathbf{q}_i| - m\right) \,\right].
\end{equation}
The classifier is thus discouraged from including out-of-box collision configurations in any predicted region, keeping the learned convex sets within physically valid regions of the configuration space. The margin $1 - m$ provides a small safety buffer from the boundary.

\noindent\textbf{False positive loss.}
The false positive loss is a hard-negative mining term that targets configurations the model currently misclassifies as inside a region.
At each training step, we identify configurations with ground-truth label $C_\text{gt} = 0$ (in collision) but predicted class probability $\geq 0.5$, and push them into a per-region ring buffer of current false positives.
At loss time, we sample $\mathbf{q}_\text{fp}$ from this buffer and apply a per-region BCE that drives the sampled points outside the region:
\begin{equation}
    \mathcal{L}_\text{fp} = \frac{1}{N} \sum_{k=1}^{N} \mathrm{BCE}\!\left(C_k(g_\theta(\mathbf{q}_\text{fp})),\; 0\right).
\end{equation}
By focusing gradient updates on regions that are currently violating the zero-false-positive criterion, this term complements the global BCE that operates over uniformly sampled configurations and accelerates convergence toward strict precision.

\subsection{Hyperparameters}
\label{app:hyperparameters}

This subsection consolidates the numerical hyperparameters used across the four environments.

\begin{table}[h]
\centering
\caption{Environment-specific hyperparameters. $N_\text{seed}$ is the visibility-guided seed budget, $N_\text{bridge}$ is the number of bridge sampling regions, $N=N_\text{seed}+N_\text{bridge}$ is the total convex set count, $B$ is the number of half-spaces per region, $m$ is the box-loss half-width, and $\lambda_\text{anc}$ is the latent-distance anchor weight.}
\label{tab:env_hyperparameters}
\small
\begin{tabular}{lrrrrrr}
\toprule
Env. & $N_\text{seed}$ & $N_\text{bridge}$ & $N$ & $B$ & $m$ & $\lambda_\text{anc}$ \\
\midrule
Nav~2D & 10 & 8 & 18 & 20 & 0.995 & 1.0 \\
Reach~6D & 2 & 1 & 3 & 100 & 0.895 & 1.0 \\
Reach~14D & 2 & 1 & 3 & 200 & 0.845 & 1.0 \\
Bimanual~14D & 2 & 1 & 3 & 100 & 0.895 & 3.0 \\
\bottomrule
\end{tabular}
\end{table}

\noindent\textbf{Loss weights.}
Each BCE term, $\mathcal{L}_\text{class}, \mathcal{L}_\text{seed}, \mathcal{L}_\text{vis}, \mathcal{L}_\text{bridge}$, applies an asymmetric weighting on outside (collision, $C_\text{gt}=0$) and inside ($C_\text{gt}=1$) samples,
\begin{equation*}
    \mathrm{BCE}(C_\text{gt},\, \hat{C}) = -\bigl[\, w_\text{out}\,(1 - C_\text{gt})\, \log(1 - \hat{C}) + C_\text{gt}\, \log \hat{C} \,\bigr],
\end{equation*}
with $w_\text{out} = 10$ biasing the BCE toward higher precision during training. The regularization term decomposes as
\begin{equation*}
    \mathcal{L}_\text{reg} = \lambda_\text{anc}\, \mathcal{L}_\text{anc} + \lambda_\text{iso}\, \mathcal{L}_\text{iso} + \lambda_\text{box}\, \mathcal{L}_\text{box} + \lambda_\text{fp}\, \mathcal{L}_\text{fp},
\end{equation*}
where each term is defined in Appendix~\ref{app:reg_losses}. Table~\ref{tab:loss_weights} lists the values of all weights in equation~(2) and in the regularization decomposition.

\begin{table}[h]
\centering
\caption{Loss weights used in equation~(2) and in the regularization decomposition. The per-environment anchor weight $\lambda_\text{anc}$ is listed in Table~\ref{tab:env_hyperparameters}.}
\label{tab:loss_weights}
\small
\begin{tabular}{ccccccccc}
\toprule
$\lambda_\text{seed}$ & $\lambda_\text{vis}$ & $\lambda_\text{bridge}$ & $\lambda_\text{reg}$ & $w_\text{out}$ & $\lambda_\text{iso}$ & $\lambda_\text{box}$ & $\lambda_\text{fp}$ & $\lambda_\text{anc}$ \\
\midrule
$0.5$ & $1.0$ & $0.5$ & $1.0$ & $10$ & $0.1$ & $1.0$ & $0.75$ & Table~\ref{tab:env_hyperparameters} \\
\bottomrule
\end{tabular}
\end{table}

\noindent\textbf{Gumbel-Softmax smoothing.}
The Gumbel-Softmax surrogates of $\min$ and $\max$ in Section~\ref{sec:lcd} approximate the maximum as
\begin{equation*}
    \widetilde{\max}_\tau(\mathbf{x}) = \sum_i x_i \cdot \mathrm{softmax}_i\!\left(x_i + \tau\, g_i\right), \quad g_i \sim \mathrm{Gumbel}(0, 1),
\end{equation*}
with $\widetilde{\min}_\tau(\mathbf{x}) = -\widetilde{\max}_\tau(-\mathbf{x})$. We set the noise scale $\tau = 0.01$, which keeps the surrogate close to the hard max/min while providing a non-zero gradient.

\noindent\textbf{Bridge sampling ratio.}
For bridge sampling (Algorithm~\ref{alg:bridge_sampling}), per-interpolation-point radii are chosen so that approximately $10\%$ of the resulting samples land in collision, providing a nearby negative signal for the bridge BCE.

\noindent\textbf{Optimizer.}
We use Adam~\citep{kingma2015adam} with $\beta_1 = 0.9$, $\beta_2 = 0.999$, no weight decay, and a constant learning rate ($2{\times}10^{-3}$ for $g_\theta$ and $0.1$ for the boundary parameters $\boldsymbol{\eta}$ and $d$). The boundary parameters are initialized with scale $0.1$.

\noindent\textbf{Hardware.}
All experiments are run on a single NVIDIA GeForce RTX 3090 with an AMD Ryzen Threadripper 3990X (64-core) CPU.

\section{Deployment Details}
\label{app:path_planning}

\subsection{Test-Time Refinement Procedure}
\label{app:refinement_detail}

Test-time refinement enforces precision\,=\,1.0 by contracting any region that has admitted a collision configuration.
We treat every false positive (a configuration the boundary classifier places \emph{inside} a region but that the collision oracle reports as unsafe) as evidence that the offending facet has drifted past the true obstacle, and we push that facet inward just enough to expel the offending point.

\paragraph{Refinement step.}
We use the notation of Section~\ref{sec:refinement}, where region $k$ is $\mathcal{P}_k = \{\mathbf{z} : \phi_{k,i}(\mathbf{z}) \ge 0, \; i = 1, \dots, B\}$ with signed distances $\phi_{k,i}(\mathbf{z}) = \boldsymbol{\eta}_{k,i}^\top \mathbf{z} + d_{k,i}$, and a false positive $\mathbf{z}_\text{fp}$ inside region $k$ satisfies $\phi_{k,i}(\mathbf{z}_\text{fp}) \ge 0$ for all $i$. Each $\mathbf{z}_\text{fp}$ is assigned to its most narrowly violated facet $i^\star(\mathbf{z}_\text{fp}) = \arg\min_i \phi_{k,i}(\mathbf{z}_\text{fp})$, giving per-facet sets $\mathcal{F}_{k,i}$, and we apply a single batched update
\[
d_{k,i} \;\leftarrow\; d_{k,i} - \bigl(\,\Delta_{k,i} + \varepsilon\,\bigr), \qquad \Delta_{k,i} = \max_{\mathbf{z}_\text{fp} \in \mathcal{F}_{k,i}} \phi_{k,i}(\mathbf{z}_\text{fp}),
\]
where $\varepsilon > 0$ is a small constant close to zero that provides a tiny safety margin. After the update, every assigned $\mathbf{z}_\text{fp}$ lies strictly outside $\mathcal{P}_k$, while configurations on the safe side of each facet remain inside.
The normals $\boldsymbol{\eta}_{k,i}$ are never modified, so each facet only translates along its own normal, meaning each region can only shrink and never grow. Because the update is independent across $(k, i)$, all facets are updated in parallel in a single batched operation.

\paragraph{Source of false positives.}
We collect false positives from two complementary sources, applied iteratively until no new violators are observed.
\begin{itemize}[leftmargin=*,topsep=2pt,itemsep=0pt]
\item \textbf{Uniform sampling sweep.}
We repeatedly draw configuration-space samples (uniform on a per-environment domain), check them against the collision oracle, and surface every sample that is classified as inside some region $\mathcal{P}_k$ but is in fact in collision.
This first pass corrects the bulk of facet overreach inherited from training.

\item \textbf{Planning-and-deploy feedback.}
After the uniform sweep converges, we run the actual GCS path planner on a held-out batch of start--goal queries and densely interpolate every returned path.
Any waypoint on a returned path that is in collision becomes a planning-time false positive and is fed back into the refinement step.
This second pass catches violators that lie on the GCS-preferred routes specifically, which uniform sampling might never hit at a useful rate in high dimensions.
\end{itemize}

Both passes only translate facets inward, so precision is monotonically non-decreasing across refinement iterations.
We stop when neither pass produces any new false positive, with a cap of $10$ outer iterations.
The safety margin is implemented as a small multiplicative offset on $\Delta_{k,i}$ ($\varepsilon \approx 10^{-6}$), and at each iteration we additionally inject $100$ Gaussian perturbations of standard deviation $5\times 10^{-3}$ around the current false positives to surface nearby violators that the planner is likely to hit at deployment.

\subsection{Start- and Goal-Snap Bridging}
\label{app:reachability}

The learned region boundaries define a strict interior region of configurations classified as safe by the model.
However, a start or goal configuration $\mathbf{q}$ may lie outside all region interiors while still being reachable from some region via a straight-line collision-free path in $\mathcal{Z}$.
To handle this, we assign $\mathbf{q}$ to the region it is most naturally connected to before invoking the planner.

More formally, given $\mathbf{z} = g_\theta(\mathbf{q})$, we project $\mathbf{z}$ onto each region $\mathcal{P}_k$ along its most violated boundary:
\begin{equation}
    i^* = \arg\min_i\, \phi_{k,i}(\mathbf{z}), \qquad
    \mathbf{z}_k^\perp = \mathbf{z} - \frac{\min(\phi_{k,i^*},\, 0)}{\|\boldsymbol{\eta}_{k,i^*}\|^2}\,\boldsymbol{\eta}_{k,i^*},
\end{equation}
and assign $\mathbf{q}$ to the nearest region $k^* = \arg\min_k \|\mathbf{z}_k^\perp - \mathbf{z}\|_2$.
Since $\phi_{k,i}(\mathbf{z}) = \boldsymbol{\eta}_{k,i}^\top \mathbf{z} + d_{k,i}$ are already available as a single matrix multiply over all $N \times B$ boundaries, the entire procedure is closed-form with $\mathcal{O}(NBn)$ cost, negligible compared to the GCS solve.

\noindent\textbf{Pool-based fallback via $k$-nearest neighbors.}
The analytic projection above only certifies that the snap point lies on a region facet.
The straight-line latent segment from $\mathbf{q}$ to the snapped configuration may still cross collision once decoded.
When this segment check fails, we fall back to a sample-based snap.
In detail, we maintain a fixed-size pool of region-inside configurations $\{\mathbf{q}_j^\text{pool}\}_{j=1}^{M}$, drawn by uniform sampling and accepted only if the boundary classifier places them strictly inside some current region $\mathcal{P}_k$.
The pool is refreshed after every test-time refinement update so it tracks the active boundaries.
Given a failing query $\mathbf{q}$, we retrieve its top-$k$ nearest pool points by configuration-space $\ell_1$ distance, linearly interpolate from $\mathbf{z} = g_\theta(\mathbf{q})$ to each candidate's latent code in $\mathcal{Z}$, decode the interpolants back to configuration space, and check that every decoded sample is collision-free.
Among the candidates passing this check, we snap $\mathbf{q}$ to the one closest in $\ell_1$ distance.
If no candidate passes the latent-linear check, we optionally retry the same procedure with interpolation directly in configuration space as a last-resort fallback, in which case the splice segment used at evaluation also switches to configuration-space linear interpolation for consistency.

\subsection{Redundant Facet Pruning}
\label{app:redundancy}

Each region
\begin{equation}
    \mathcal{P}_k = \{ \mathbf{z} \in \mathcal{Z} : \phi_{k,i}(\mathbf{z}) \ge 0,\; i = 1, \dots, B \}, \quad \phi_{k,i}(\mathbf{z}) = \boldsymbol{\eta}_{k,i}^\top \mathbf{z} + d_{k,i},
\end{equation}
often contains redundant facets whose removal leaves $\mathcal{P}_k$ unchanged.
These facets inflate the pairwise region-intersection checks used to build the GCS adjacency graph and the per-vertex convex programs solved during GCS optimization.
We prune them once per active boundary set $\mathcal{H}$ and cache the result, so the cost is paid only at planner-time refinement updates and amortized over all evaluation queries.

\noindent\textbf{Stage 1: Gram-matrix prefilter.}
For each region, let $\hat{\boldsymbol{\eta}}_{k,i} = \boldsymbol{\eta}_{k,i} / \|\boldsymbol{\eta}_{k,i}\|$ and $\hat{d}_{k,i} = d_{k,i} / \|\boldsymbol{\eta}_{k,i}\|$.
We form $G_{ij} = \hat{\boldsymbol{\eta}}_{k,i}^\top \hat{\boldsymbol{\eta}}_{k,j}$ and, for every pair with $G_{ij} > 1 - \tau_\text{cos}$, drop the facet with the larger $\hat{d}$.
Two co-oriented parallel half-spaces are strictly nested, so dropping the larger-$\hat{d}$ one preserves $\mathcal{P}_k$ exactly.
We set $\tau_\text{cos} = 10^{-10}$ so only floating-point duplicates are merged, and defer near-parallel but distinct facets to Stage~2.

\noindent\textbf{Stage 2: Per-facet LP check.}
Surviving facets are tested sequentially with the linear program
\begin{equation}
    \phi_i^\star = \min_{\mathbf{z}}\; \phi_{k,i}(\mathbf{z}) \quad \text{s.t.}\quad \phi_{k,j}(\mathbf{z}) \ge 0 \;\;\forall j \in K \setminus \{i\},
    \label{eq:lp-redundancy}
\end{equation}
where $K$ is the current active index set.
Facet $i$ is dropped when the LP returns a bounded optimum with $\phi_i^\star \ge -\tau_\text{LP}$ ($\tau_\text{LP} = 10^{-7}$).
Already-dropped facets are excluded from $K$ before later facets are tested, so no redundant facet is certified using another.
Unbounded or numerically problematic LPs conservatively retain the facet.
The procedure preserves the shape of every $\mathcal{P}_k$ to within floating-point noise and only reduces the description size handed to the GCS solver.

\section{Additional Experimental Details} \label{app:details}

\subsection{Coverage Metric Definitions}
\label{app:coverage_metric}

This appendix gives the full formal definitions of the three coverage metrics $\text{Cvg}_q$, $\text{Cvg}_p$, $\text{Cvg}_r$ used in Section~\ref{sec:result_repr} and Table~\ref{tab:repr_flat}.
Let $\Omega = \hat{\mathcal{Q}}_\text{free}^{(\text{island})} \cap \mathcal{Q}_\text{free}$ be the ground-truth collision-free samples lying in the largest connected component of the predicted region, so that only configurations the planner can actually route through are counted.
The three metrics measure coverage in configuration space, end-effector position space, and end-effector pose space respectively, defined as
\begin{equation}
    \text{Cvg}_q = \frac{|\Omega|}{|\mathcal{Q}_\text{free}|}, \quad
    \text{Cvg}_p = \frac{|\mathrm{Vox}(\mathrm{FK}_p(\Omega))|}{|\mathrm{Vox}(\mathrm{FK}_p(\mathcal{Q}_\text{free}))|}, \quad
    \text{Cvg}_r = \frac{|\mathrm{Vox}(\mathrm{FK}_r(\Omega))|}{|\mathrm{Vox}(\mathrm{FK}_r(\mathcal{Q}_\text{free}))|},
\end{equation}
where $\mathrm{FK}_p(\cdot)$ maps a configuration to its end-effector position, $\mathrm{FK}_r(\cdot)$ concatenates the positions of multiple rigidly-attached keypoints (capturing both position and orientation implicitly), and $\mathrm{Vox}(\cdot)$ voxelizes the workspace.

\noindent\textbf{Keypoints and voxel sizes.}
For $\mathrm{FK}_r$, we place keypoints on the last two link bodies of each arm, namely the end-effector link and the link immediately preceding it. For the single-arm reachers (Reach~6D, Reach~14D), the 2 keypoints are concatenated into a single position vector before voxelization. For Bimanual~14D, a combined $12$-D voxel grid (4 keypoints $\times$ 3 dims) would be too sparse at any reasonable resolution, making coverage estimates unreliable. We therefore compute $\text{Cvg}_r$ per arm using each arm's 2 keypoints concatenated into a $6$-D vector, and average the two per-arm values. The $\mathrm{Vox}(\cdot)$ voxel size is $0.2$ for Nav~2D, $0.02$ for the reacher environments, and $0.05$ for Bimanual~14D.

\subsection{Planning Evaluation Setup}
\label{app:planning_eval_setup}

For downstream planning evaluation, we use a Drake-based GCS planner constructed on top of the learned collision-free region decomposition.
We report path-planning success rate (SR) and planning time.
A planning trial is counted as successful if the forward-kinematics keypoint error between the target configuration and the final planned configuration is below a predefined threshold.
Unless otherwise noted, all methods are evaluated under the same environment setting and number of convex sets so that the results directly reflect differences in representation quality and planning utility.

\subsection{Path Planning Dataset}
\label{app:planning_dataset}

For each environment, the evaluation set consists of $1{,}000$ start--goal configuration pairs $(\mathbf{q}_\text{s}, \mathbf{q}_\text{g})$.
Both endpoints are sampled uniformly from $\mathcal{Q}_\text{free}$, and a pair is rejected if the straight-line segment between them lies entirely inside $\mathcal{Q}_\text{free}$, since such trivial pairs do not exercise the region representation.
This procedure is repeated until $1{,}000$ accepted pairs are collected.
The one exception is IRIS on Bimanual~14D, where each per-pair GCS solve takes minutes due to its $90$-region decomposition in $14$ dimensions, so we evaluate it on $50$ pairs drawn from the same distribution instead.

\subsection{Baseline Hyperparameter Tuning}
\label{app:baseline_tuning}

The sampling-based baselines (RRT, PRM, cuRobo) approximate the world geometry with collision primitives whose effective shape depends on each environment's robot model.
Because the modeled obstacle and robot geometries differ across our environments (a 2D point robot in Nav~2D, planar manipulators in Reach~6D and Reach~14D, and a 3D bimanual manipulator in Bimanual~14D), a single set of collision-margin hyperparameters does not transfer across environments.
We therefore tune each baseline per-environment for a fair comparison.

\noindent\textbf{RRT-Connect and PRM (OMPL, capsule primitives).}
Both planners run in OMPL with a MuJoCo state validity checker. Each robot link and obstacle is represented by its native capsule shell, so the underlying primitive is the capsule.
Analogously to the cuRobo sphere-padding sweep, we sweep a per-environment collision margin $m$ that inflates the non-robot collision geom radii while leaving the robot capsules untouched.
The swept values are $m \in \{0.001,\,0.002,\,0.005,\,0.010,\,0.020,\,0.050,\,0.100\}$\,m, and for Table~\ref{tab:planning_eval} we pick whichever value yields the highest post-validation SR per environment.

\noindent\textbf{cuRobo (sphere primitives).}
cuRobo's collision oracle expects a sphere-cloud representation of the robot, so we transcribe each capsule of the source robot model into a uniformly spaced set of spheres that covers the capsule's swept volume exactly.
For a capsule of length \(L\) and radius \(r\), we place \(N=\max(3,\lceil L/r\rceil + 1)\) spheres along its axis with radius \(R=\sqrt{r^2 + (L/(2(N-1)))^2}\). This is the smallest \(N\) for which a uniformly spaced cover has no axial gap, and the chosen \(R\) keeps the sphere envelope tangent to the original capsule, so cuRobo's robot-world collision boundary matches the MuJoCo geometry without arbitrary inflation.
Self-collision ignore pairs are seeded from kinematic adjacency.
Where thick-link envelopes always overlap with non-adjacent neighbors, we add link pairs that overlap in \(\geq 95\%\) of \(500\) randomly sampled configurations to the ignore list (pairs that overlap only sometimes are kept since they encode real near-collisions).
We verify each resulting collision oracle against MuJoCo on \(500\) held-out random configurations and require agreement \(\geq 95\%\) with false-negative rate \(< 0.5\%\) before running the planning evaluation.
We then sweep the robot-side collision sphere padding per environment with cuRobo's graph-search variant and reuse the best value for the trajectory-optimization variant.
The padding grids are $\{0.015,\,0.020,\,0.025,\,0.030\}$\,m for Nav~2D, $\{0.001, 0.002, \dots, 0.010\}$\,m (1\,mm step) for Reach~6D and Reach~14D, and $\{0,\,0.005, 0.010, \dots, 0.045\}$\,m (5\,mm step) for Bimanual~14D. Per environment we pick the padding with the highest SR (ties broken by lowest solve time), giving $\{0.025, 0.002, 0.003, 0.010\}$\,m for the four environments respectively, and reuse that value for the trajectory variant.

In contrast, ILD does not reduce the environment to either capsule shells or sphere clouds. Every obstacle's full mesh is presented to the boundary-learning sampler, and the learned safe region is what GCS plans on top of.
The SR advantage at higher dimensions reported in \Cref{tab:planning_eval} reflects this primitive-free treatment of the obstacle geometry.

\section{Geometric Interpretation of the Joint Objective}
\label{app:bce_gradient}

We show that the BCE objective in Section~\ref{sec:lcd} geometrically encourages $g_\theta$ to warp non-convex parts of $\mathcal{Q}_\text{free}$ into convex-friendly shapes so that the polytopes can match safe and unsafe regions. The derivation below makes precise how each gradient step acts on the signed distance from a sample to its closest polytope facet, and shows that the effect monotonically improves the sample's classification.

\paragraph{Setup.}
With $\mathbf{z} = g_\theta(\mathbf{q})$ and $s := \max_k \varphi_k(\mathbf{z})$, the BCE loss for sample $\mathbf{q}$ is
\[
\mathcal{L}_\text{BCE}(\mathbf{q}) \;=\; -\bigl[\, C_\text{gt}\, \log C(\mathbf{z}) + (1 - C_\text{gt})\, \log(1 - C(\mathbf{z})) \,\bigr], \qquad C(\mathbf{z}) = \sigma(s),
\]
with $C_\text{gt} \in \{0, 1\}$ the ground-truth label at $\mathbf{q}$.
In the hard-max/min limit, $s$ is realised at the argmax region $k^* = \arg\max_k \varphi_k(\mathbf{z})$ and its argmin facet $i^* = \arg\min_i \phi_{k^*, i}(\mathbf{z})$, giving the local gradient $\partial s / \partial \mathbf{z} = \boldsymbol{\eta}_{k^*, i^*}$.

\paragraph{Gradient identity.}
The standard BCE\,+\,sigmoid identity yields $\partial \mathcal{L}_\text{BCE}/\partial s = C(\mathbf{z}) - C_\text{gt}$. Chaining through $\mathbf{z} = g_\theta(\mathbf{q})$,
\begin{equation}
    \nabla_\theta \mathcal{L}_\text{BCE} \;=\; \bigl(\, C(\mathbf{z}) - C_\text{gt} \,\bigr)\, \boldsymbol{\eta}_{k^*, i^*}^\top\, \nabla_\theta\, g_\theta(\mathbf{q}).
\end{equation}

\paragraph{Effect on the signed distance.}
A gradient step $\Delta\theta = -\alpha \nabla_\theta \mathcal{L}_\text{BCE}$ changes the signed distance $\phi_{k^*, i^*}(g_\theta(\mathbf{q}))$ from the sample to its closest facet by
\begin{equation}
    \Delta \phi_{k^*, i^*}
    \;\approx\; \underbrace{\boldsymbol{\eta}_{k^*, i^*}^\top \nabla_\theta g_\theta(\mathbf{q})}_{\partial \phi_{k^*, i^*} / \partial \theta} \cdot \Delta\theta
    \;=\; -\alpha\, \bigl(\, C(\mathbf{z}) - C_\text{gt} \,\bigr)\, \bigl\|\boldsymbol{\eta}_{k^*, i^*}^\top \nabla_\theta g_\theta(\mathbf{q})\bigr\|^2,
\end{equation}
whose non-negative norm factor implies $\mathrm{sign}(\Delta \phi_{k^*, i^*}) = -\mathrm{sign}(C(\mathbf{z}) - C_\text{gt})$, hence
\begin{equation}
    \underbrace{\Delta \phi_{k^*, i^*} \le 0 \;\Longleftrightarrow\; C(\mathbf{z}) > C_\text{gt}}_{\text{false positive}},
    \qquad
    \underbrace{\Delta \phi_{k^*, i^*} \ge 0 \;\Longleftrightarrow\; C(\mathbf{z}) < C_\text{gt}}_{\text{false negative}}.
\end{equation}
A false positive is pushed across facet $i^*$ out of polytope $k^*$, while a false negative is pulled deeper inside.

\paragraph{Connection to test-time refinement (single-step).}
The bias gradient reduces to test-time refinement under a few explicit assumptions. In the hard limit $s = \phi_{k^*, i^*}(\mathbf{z})$, so $\partial s / \partial d_{k^*, i^*} = 1$ and
\begin{equation}
    \frac{\partial \mathcal{L}_\text{BCE}}{\partial d_{k^*, i^*}} = C(\mathbf{z}) - C_\text{gt},
    \qquad
    \Delta d_{k^*, i^*} = -\alpha\,\bigl(C(\mathbf{z}) - C_\text{gt}\bigr).
\end{equation}
For a false positive ($C_\text{gt} = 0$) this decreases $d_{k^*, i^*}$, translating facet $i^*$ inward to expel the sample, the same direction as the test-time contraction $d_{k^*, i^*} \leftarrow d_{k^*, i^*} - \Delta_{k^*, i^*}$ (Section~\ref{sec:refinement}). The two updates are not identical, but the gap closes under a few mild assumptions that make the train-time gradient collapse onto the test-time rule.
\begin{itemize}[leftmargin=*,topsep=2pt,itemsep=1pt]
\item \textbf{Frozen geometry.} Test-time refinement holds $g_\theta$ and the normals $\boldsymbol{\eta}$ fixed and moves only the bias, whereas training co-adapts all of them.
\item \textbf{One-sided contraction.} The train-time residual $C(\mathbf{z}) - C_\text{gt}$ is two-sided, contracting a facet on a false positive ($C_\text{gt} = 0$) and expanding it on a false negative ($C_\text{gt} = 1$). Test-time refinement keeps only the false-positive branch, so boundaries can only shrink and precision is monotone.
\item \textbf{Raw margin in place of the squashed residual.} The gradient step scales with the sigmoid-squashed residual $C(\mathbf{z}) - C_\text{gt} = \sigma(\phi_{k^*, i^*}(\mathbf{z}))$, a bounded probability in $(0,1)$ applied repeatedly with learning rate $\alpha$. Test-time refinement instead steps by the raw signed-distance violation, setting $\Delta_{k^*, i^*} = \max_{\mathbf{z}_\text{fp}} \phi_{k^*, i^*}(\mathbf{z}_\text{fp})$ and adding a small $\varepsilon$ so that every assigned false positive lands strictly outside the contracted facet in a single shot.
\end{itemize}
Under these assumptions the two updates coincide in direction and target facet. They differ only in whether the contraction is the soft, probability-weighted, repeated gradient step or the exact, geometric, single-shot margin. Because the test-time step is one-sided, iterating it like the gradient would over-contract, so it uses the minimal exact amount $\Delta_{k^*, i^*} + \varepsilon$ rather than a repeated nudge, whereas during training the opposing false-negative gradients from safe samples on the inner side of the facet keep the boundary in balance.

\paragraph{Connection to test-time refinement (fixed-point).}
Beyond a single step, training and test-time refinement share a fixed point in the hard-sigmoid limit ($\beta \to \infty$), where the BCE gradient vanishes and the refinement update leaves $d$ unchanged. Project each sample assigned to facet $i^*$ of region $k^*$ onto the facet normal, $t_n := \boldsymbol{\eta}_{k^*, i^*}^\top \mathbf{z}_n$, so $\phi_{k^*, i^*}(\mathbf{z}_n) = t_n + d$ with $d \equiv d_{k^*, i^*}$ the only free variable. The classifier on this facet is $\sigma\!\big(\beta(t_n + d)\big)$, where $\beta = 1$ recovers the standard sigmoid and $\beta \to \infty$ gives the hard step. With $\partial s / \partial d = 1$, stationarity of the two-sided BCE in $d$ gives
\begin{equation}
    \sum_n \bigl(\sigma(\beta(t_n + d)) - C_{\text{gt}, n}\bigr) = 0
    \;\iff\;
    \underbrace{\sum_{n : C_{\text{gt}, n} = 0} \sigma(\beta(t_n + d))}_{\text{soft FP mass}}
    \;=\;
    \underbrace{\sum_{n : C_{\text{gt}, n} = 1} \sigma(-\beta(t_n + d))}_{\text{soft FN mass}},
\end{equation}
so the boundary settles where the predicted-inside mass of collision samples balances the predicted-outside mass of free samples.

Take the hard-sigmoid limit $\beta \to \infty$. Then $\sigma(\beta(t + d)) \to \mathbf{1}[t > -d]$ and the soft masses collapse into integer counts of mis-classified samples,
\begin{equation}
    N_\text{FP}(d) \;:=\; |\{ n : C_{\text{gt}, n} = 0,\; t_n > -d \}|,
    \qquad
    N_\text{FN}(d) \;:=\; |\{ n : C_{\text{gt}, n} = 1,\; t_n < -d \}|,
\end{equation}
and the balance reduces to $N_\text{FP}(d) = N_\text{FN}(d)$. Dropping the FN term under the one-sided restriction leaves the constraint $t_n + d \le 0$ for every collision $n$, the requirement that no collision sample is placed strictly inside the facet. The minimal-contraction choice, the largest $d$ satisfying it, gives
\begin{equation}
    d^\star \;=\; -\max_{n : C_{\text{gt}, n} = 0} t_n.
\end{equation}
Substituting $\Delta_{k^*, i^*} = \max_{n : C_{\text{gt}, n} = 0}(t_n + d_\text{old})$ into the test-time refinement update,
\begin{equation}
    d_\text{new}
    \;=\; d_\text{old} \;-\; \Delta_{k^*, i^*}
    \;=\; -\max_{n : C_{\text{gt}, n} = 0} t_n
    \;=\; d^\star,
\end{equation}
so a single refinement step lands exactly on the fixed point of the hard-sigmoid one-sided problem. Removing either limit breaks the identity. Keeping the soft sigmoid replaces the count with a probability mass that never hits zero, and keeping the FN term adds a competing pull that settles the boundary strictly inside the gap between collision and free samples.

\end{document}